\documentclass[10pt,twocolumn,letterpaper]{article}

\usepackage[pagenumbers]{iccv} % To force page numbers, e.g. for an arXiv version
\usepackage{algorithmic}
\usepackage{algorithm}
\usepackage{multirow}

\definecolor{iccvblue}{rgb}{0.21,0.49,0.74}
\usepackage[pagebackref,breaklinks,colorlinks,allcolors=iccvblue]{hyperref}

\def\paperID{7944} % *** Enter the Paper ID here
\def\confName{ICCV}
\def\confYear{2025}

\title{Unveiling Hidden Vulnerabilities in Digital Human Generation via Adversarial Attacks}

\author{Zhiying Li$^{a}$\thanks{These authors contributed equally.}, Yeying Jin$^{d}$ \footnotemark[1], Fan Shen$^e$, Zhi Liu$^a$, Weibin Chen$^a$, Pengju Zhang$^f$, Xiaomei Zhang$^g$, \\
Boyu Chen$^h$, Michael Shen$^i$, Kejian Wu$^j$, Zhaoxin Fan$^{b,c,{\dagger}}$, Jin Dong$^{k,{\dagger}}$\\
$^a$\textit{College of Cyber Security, Jinan University}\\
$^b$\textit{Beijing Advanced Innovation Center for Future Blockchain and Privacy Computing, }\\
\textit{School of Artificial Intelligence, Beihang University}\\
$^c$\textit{Hangzhou International Innovation Institute, Beihang University}\\
$^d$\textit{Department of Electrical and Computer Engineering, National University of Singapore}\\
$^e$\textit{Department of Electrical and Computer Science, University of Pittsburgh}\\
$^f$\textit{State Key Laboratory of Multimodal Artificial Intelligence, }\\
\textit{Institute of Automation, Chinese Academy of Science}\\
$^g$\textit{Institute of Automation, Chinese Academy of Science}\\
$^h$\textit{University College London}, $^i$\textit{Mingdu Tech}, $^j$\textit{Xreal}\\
$^k$\textit{Beijing Academy of Blockchain and Edge Computing} 
}

\begin{document}
\maketitle

\begin{abstract}
Expressive human pose and shape estimation (EHPS) is crucial for digital human generation, especially in applications like live streaming. While existing research primarily focuses on reducing estimation errors, it largely neglects robustness and security aspects, leaving these systems vulnerable to adversarial attacks. To address this significant challenge, we propose the \textbf{Tangible Attack (TBA)}, a novel framework designed to generate adversarial examples capable of effectively compromising any digital human generation model. Our approach introduces a \textbf{Dual Heterogeneous Noise Generator (DHNG)}, which leverages Variational Autoencoders (VAE) and ControlNet to produce diverse, targeted noise tailored to the original image features. Additionally, we design a custom \textbf{adversarial loss function} to optimize the noise, ensuring both high controllability and potent disruption. By iteratively refining the adversarial sample through multi-gradient signals from both the noise and the state-of-the-art EHPS model, TBA substantially improves the effectiveness of adversarial attacks. Extensive experiments demonstrate TBA's superiority, achieving a remarkable 41.0\% increase in estimation error, with an average improvement of approximately 17.0\%. These findings expose significant security vulnerabilities in current EHPS models and highlight the need for stronger defenses in digital human generation systems.
\end{abstract}

%==============================================================
\section{Introduction}

\begin{figure}[t]
    \centering
    \includegraphics[width=0.45\textwidth]{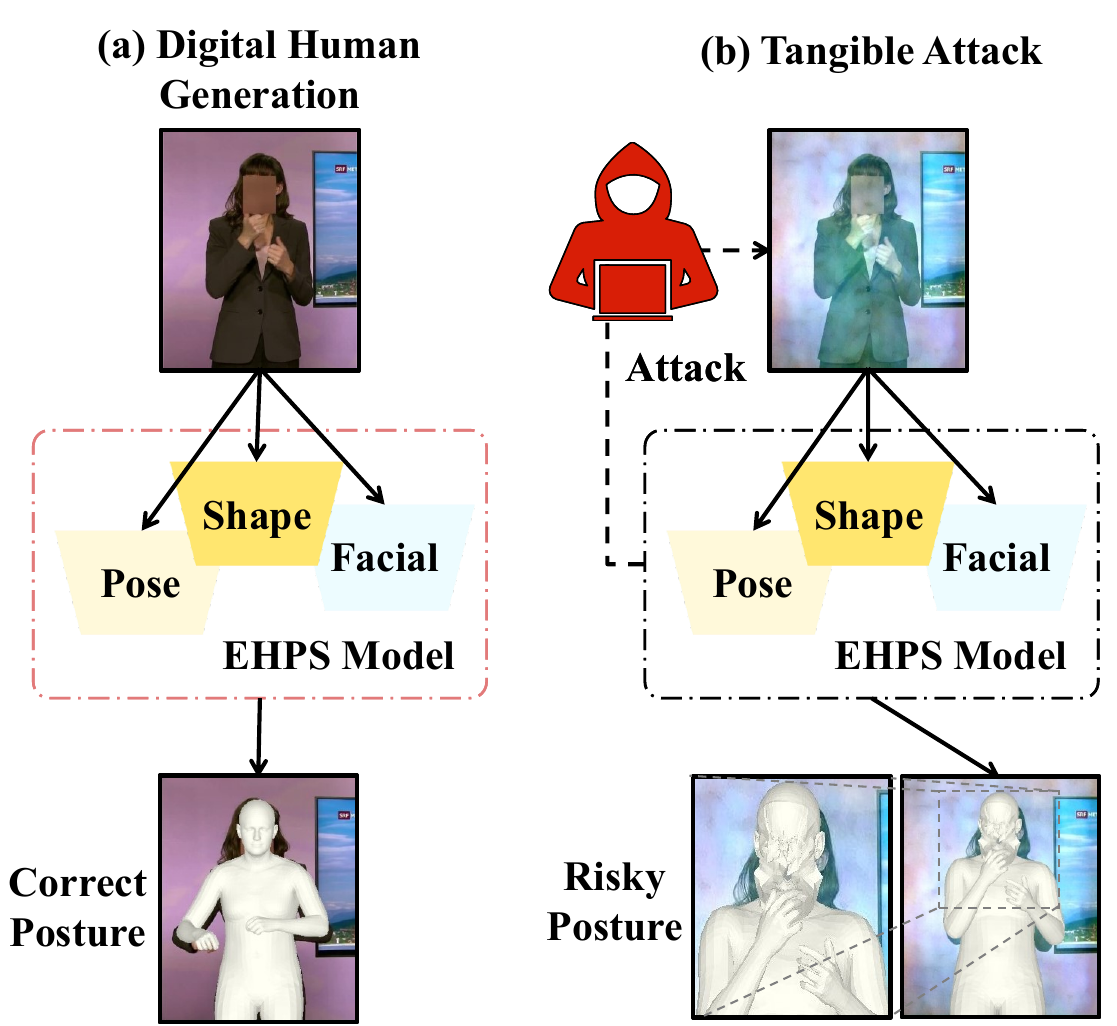}
    \caption{Under normal conditions, a clean image generates a realistic digital human with the correct posture. During an attack, adversarial samples cause significant deviations in the model’s output, resulting in risky posture.}
    % \caption{Illustration of an attack on a digital human generation model. Under normal conditions, a clean image generates a realistic digital human. During an attack, adversarial samples cause significant deviations in the model’s output.}
    \label{motivation}
\end{figure}

Expressive Human Pose and Shape Estimation (EHPS) \cite{loper2015smpl, pavlakos2019expressive, hong2022avatarclip, hong2021garment4d} from monocular images or videos is a critical task in digital human generation, with applications spanning live streaming, virtual reality (VR), and gaming. In these applications, EHPS is essential for creating realistic, interactive digital humans that can mimic human body movement and expressions in real time. However, the vulnerability of these systems to adversarial attacks poses significant risks \cite{diao2024understanding, chen2024towards, schmalfuss2023distracting, zhou2024robustness}, as shown in Fig. \ref{motivation}. For instance, in live streaming, adversarial samples could disrupt the digital human’s behavior, leading to broadcast incidents or reputational damage. In VR environments, such attacks could be exploited by malicious actors to manipulate avatars or even simulate harmful scenarios, such as enabling acts of virtual terrorism. These potential threats highlight the urgent need for robustness and security in EHPS systems to prevent real-world consequences.

Most existing research on EHPS focuses on improving estimation accuracy by reducing errors in body \cite{zhou2021monocular, zhang2023pymaf}, face, and hand pose estimation \cite{jin2020whole, moon2022accurate, lin2023one, cai2024smpler}. This is typically achieved by leveraging parameterized human models such as SMPL-X \cite{pavlakos2019expressive} or SMPLer-X \cite{cai2024smpler}, which use deep neural networks to capture the intricate details of human shape and motion. While these models have shown impressive performance, their robustness against adversarial attacks has been largely overlooked. As a result, EHPS models are susceptible to adversarial examples that can significantly distort their estimations, making digital humans uncontrollable and unreliable in real-world deployments. Despite widespread advances in adversarial defense across other domains, such as image classification \cite{giulivi2023adversarial, chen2024data, xiao2023revisiting} and object detection \cite{liu2025radap, pintor2023imagenet, dai2022deep}, the security of EHPS remains underexplored.
% \cite{goodfellow2014explaining, kurakin2018adversarial, moosavi2016deepfool, zeng2019adversarial, madry2017towards, wang2021admix, sriramanan2020guided}

To address this critical gap, we propose the \textbf{Tangible Attack (TBA)} framework, a novel adversarial attack method specifically designed to target EHPS systems. The TBA framework generates adversarial samples capable of launching effective attacks on any digital human generation model. Instead of relying on a single noise generation process, our approach introduces a \textbf{Dual Heterogeneous Noise Generator (DHNG)}, which combines two distinct noise generation techniques. One is based on a Variational Autoencoder (VAE) \cite{mansour2021unsupervised}, which manipulates latent space representations to introduce variations, while the other uses ControlNet \cite{zhang2023adding}, a conditional control mechanism that refines the noise based on specific visual features. Together, these components allow DHNG to generate diverse and potent adversarial perturbations that exploit the vulnerabilities of EHPS models. Additionally, we introduce a novel \textbf{adversarial loss function} specifically designed to optimize the generated noise, ensuring it is both destructive to the model's performance and controllable in terms of intensity. To enhance the effectiveness of the attack, we employ a multi-gradient approach, inspired by the Projected Gradient Descent (PGD) method \cite{madry2018towards}, which iteratively refines the noise by leveraging gradients from both the input images and the target model (e.g., SMPLer-X). This iterative process creates more effective perturbations that target multiple feature domains within the model, maximizing the attack's success.

We validate the effectiveness of the TBA framework through extensive experiments against multiple state-of-the-art EHPS models \cite{moon2022accurate, lin2023one, cai2024smpler}. Our results demonstrate that TBA significantly increases the estimation error in these models, exposing critical security vulnerabilities that could be exploited in real-world applications. To the best of our knowledge, this is the first study to systematically investigate the robustness of EHPS through adversarial attacks.

Our main contributions are as follows:
\begin{itemize}
    \item We propose the \textbf{Tangible Attack (TBA)}, the first adversarial attack framework specifically designed to target EHPS, capable of generating adversarial samples that effectively compromise various digital human generation models.
    
    \item We introduce the \textbf{Dual Heterogeneous Noise Generator (DHNG)}, which combines Variational Autoencoder (VAE) and ControlNet to produce diverse and effective adversarial perturbations by capturing critical features from different perspectives.
    
    \item We design a novel \textbf{adversarial loss function} and employ a multi-gradient approach to iteratively optimize the adversarial noise, enhancing the attack's potency while maintaining control over its intensity.
\end{itemize}

%==============================================================
\section{Related Work}
% \cite{wang2020sequential, li2021hybrik, pavlakos2018learning, zou2021eventhpe, pang2022benchmarking, shen2023global}
\subsection{Expressive Human Pose and Shape Estimation}
Expressive Human Pose and Shape Estimation (EHPS) has a wide range of applications, from virtual reality to live streaming and animation, making it a critical task in digital human generation \cite{li2021hybrik, zou2021eventhpe, pang2022benchmarking, shen2023global}. Significant progress has been made in improving the accuracy and realism of EHPS models in recent years. Pavlakos et al. \cite{pavlakos2019expressive} introduced SMPL-X, a breakthrough in human pose estimation by incorporating articulated hands and expressive faces into a unified 3D parametric model, thereby enhancing the representation of human behavior from a single image. Moon et al. \cite{moon2022accurate} developed Hand4Whole, which leverages MCP joint features to precisely predict 3D wrist rotations while excluding body features for accurate 3D finger pose estimation. This approach significantly improved the accuracy of 3D hand pose estimation. Lin et al. \cite{lin2023one} proposed OSX, a one-stage framework that uses a Component Aware Transformer to recover whole-body meshes, simplifying the process by avoiding post-processing while achieving natural mesh predictions. Cai et al. \cite{cai2024smpler} introduced SMPLer-X, a generalist foundation model trained on diverse datasets, which demonstrated strong performance and transferability in expressive human pose and shape estimation across various benchmarks.

While these advancements have greatly contributed to the field of digital human generation, they primarily focus on reducing estimation errors. Whether these models are robust and can withstand attacks remains unknown. In this paper, we mainly investigate the robustness of these models using Adversarial Attack

% \cite{huang2021wars, guo2019simple, shen2023global, hu2022adversarial, huang2020universal, zhang2024comprehensive}
\subsection{Adversarial Attack to Deep Learning Models}
Adversarial attacks are a common method for testing the robustness of machine learning models, particularly deep neural networks (DNNs), against malicious inputs \cite{huang2021wars, shen2023global, huang2020universal, zhang2024comprehensive}. These attacks exploit vulnerabilities by introducing subtle perturbations to input data, which lead to incorrect predictions even when the perturbed data appears nearly identical to the original. Madry et al.~\cite{madry2018towards} highlight the inherent vulnerability of DNNs to adversarial examples and propose a robust optimization framework to improve adversarial robustness. Their work offers a unified perspective on existing methods and introduces reliable techniques for training networks to resist a broad spectrum of adversarial attacks. Liu et al. \cite{diao2024understanding} develop CIASA, a targeted adversarial attack on skeleton-based human action recognition using graph convolutional networks. CIASA perturbs joint locations while maintaining temporal coherence and spatial integrity, successfully deceiving state-of-the-art models with high confidence, demonstrating the vulnerability of DNNs to such attacks. Similarly, Wang et al. \cite{liu2020adversarial} introduce SMART, an adversarial attack method that targets the robustness of skeleton-based action recognition systems. Their innovation lies in the use of a perceptual loss function that captures motion dynamics, allowing for larger, yet imperceptible, perturbations that challenge the integrity of top-performing models. Recently, Zhang et al. \cite{liu2023local} devise a constrained optimization technique for imperceptible adversarial attacks on human pose estimation. By employing gradient refinement and selective pixel perturbation, their method successfully compromises leading human pose estimation (HPE) models with minimal visual alterations.

Despite these important advancements, the domain of digital human generation, particularly in the context of EHPS, remains underexplored in terms of adversarial robustness. In this paper, we investigate the robustness of EHPS—an essential task in digital human generation—against adversarial samples, revealing its underlying security vulnerabilities.

%==========================================================

\section{Preliminaries}

\subsection{Expressive Human Pose and Shape Estimation}
Expressive Human Pose and Shape (EHPS) estimation aims to generate detailed 3D representations of the human body, face, and hands from monocular images. This task is critical for applications in animation, virtual reality, and human-computer interaction. Formally, the ground truth pose parameters are defined as $\delta \in \Omega^{53 \times 3}$, comprising body poses $\delta_{\text{body}} \in \Omega^{22 \times 3}$, left-hand poses $\delta_{\text{lhand}} \in \Omega^{15 \times 3}$, right-hand poses $\delta_{\text{rhand}} \in \Omega^{15 \times 3}$, and jaw poses $\delta_{\text{jaw}} \in \Omega^{1 \times 3}$. Additionally, shape parameters $\rho \in \Omega^{10}$ and facial expression parameters $\varphi \in \Omega^{10}$ capture variations in individual anatomy and expression.

Given an EHPS model $\mathcal{F}$ and a monocular image $x$, the task is to estimate pose, shape, and expression parameters $\left \{ \delta , \rho, \varphi \right \}$:
\begin{gather}
    \mathcal{F}(x) = \left \{ \hat{\delta} , \hat{\rho}, \hat{\varphi} \right \}, \\
    \min_{\hat{\delta}, \hat{\rho}, \hat{\varphi}} \left \{ \|\delta - \hat{\delta}\| + \|\rho - \hat{\rho}\| + \|\varphi - \hat{\varphi}\| \right \}.
\end{gather}

The optimal model $\mathcal{F}$ generates parameters that approximate ground truth on unseen images. The 3D joints are then derived using $\mathcal{G}(\mathcal{S}(\delta, \rho, \varphi))$, where $\mathcal{S}$ is the SMPL-X function and $\mathcal{G}$ is the joint regressor.

\subsection{The Main Pipeline of Adversarial Attack}
Adversarial attacks introduce subtle, carefully designed perturbations to disrupt the performance of deep neural networks (DNNs), posing significant security risks across various applications. These attacks typically generate imperceptible modifications to the input, leading models to produce incorrect outputs. Formally, given a model $f$, an image $x$, and its ground truth label $y$, the adversarial attack can be formulated as:
\begin{equation}
    f(\hat{x}) \ne y \quad \text{s.t.} \quad \|\hat{x} - x\| < \varepsilon,
\end{equation}

where $\varepsilon$ denotes the maximum allowable perturbation. 

%=================================================
\section{Approach}
In this section, we first formalize the problem and then present the TBA framework in detail.

\subsection{Problem Formulation}
Given an EHPS model $\mathcal{F}$ for digital human generation and a clean input image $\mathcal{I}^{c}$, an adversarial sample $\mathcal{I}^{adv}$ can significantly increase the estimation error of $\mathcal{F}$, as follows:
\begin{gather}
    \mathcal{F}(\mathcal{I}^{c}) = \left \{ \delta^{c} \in \Omega ^{53 \times 3}, \rho^{c}  \in \Omega ^{10}, \varphi^{c} \in \Omega ^{10} \right \}, \\
    \mathcal{F}(\mathcal{I}^{adv}) = \left \{ \delta^{adv} \in \Omega ^{53 \times 3}, \rho^{adv}  \in \Omega ^{10}, \varphi^{adv} \in \Omega ^{10} \right \}, \\
    \delta^{adv} \ne \delta^{c}, \rho^{adv} \ne \rho^{c}, \varphi^{adv} \ne \varphi^{c} \quad \text{s.t.} \quad \left \| \mathcal{I}_c - \mathcal{I}_{adv} \right \| < \varepsilon ,
\end{gather}

where $\left \| \cdot \right \|$ is a distance metric quantifying the similarity between the clean image $\mathcal{I}^{c}$ and the adversarial image $\mathcal{I}^{adv}$, and $\varepsilon$ is a threshold ensuring the perturbation remains imperceptible. The 3D parametric model SMPL-X \cite{pavlakos2019expressive} is employed to represent the body, face, and hands using a set of pose, shape, and expression parameters.

\begin{figure*}[t]
    \centering
    \includegraphics[width=0.95\textwidth]{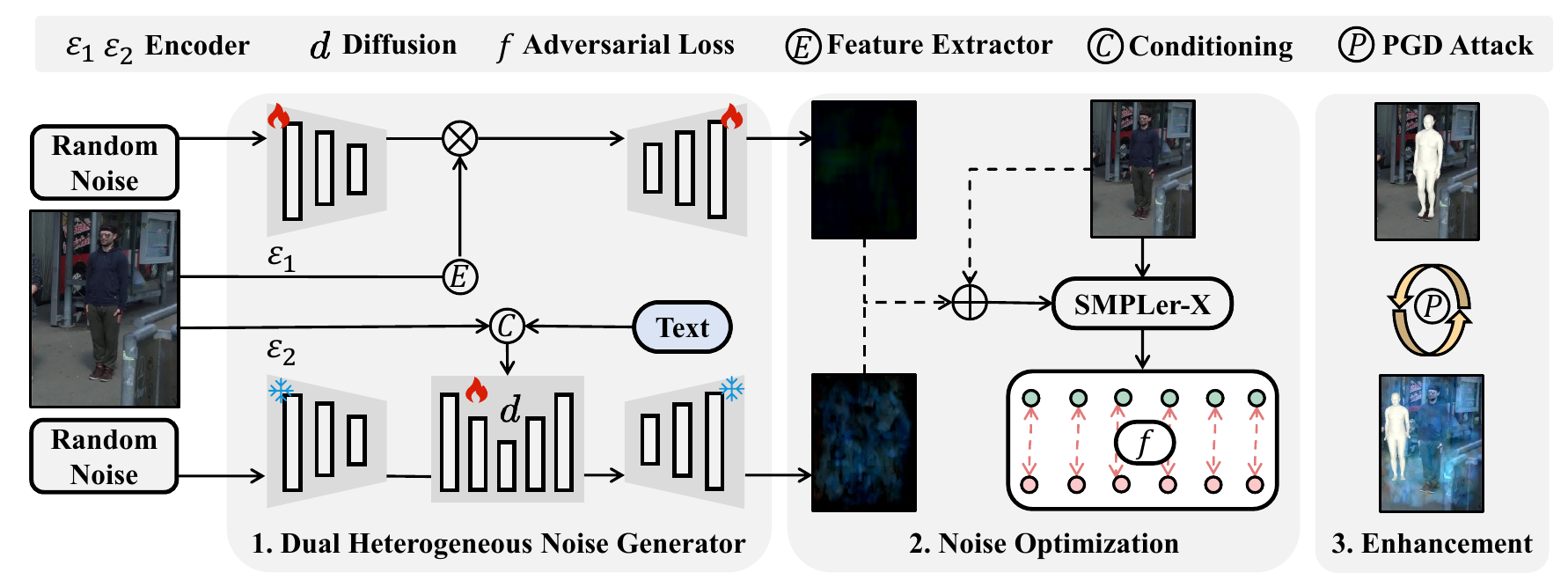}
    \caption{The overflow of the TBA framework is as follows: 1) Dual Heterogeneous Noise Generators: Utilizing a combination of VAE and ControlNet, this component generates diverse and targeted noise tailored to the characteristics of the original image. 2) Noise Optimization: The generated noise is refined through the application of a novel adversarial loss function and the gradients of the SMPLer-X model. 3) Enhancement: The attack efficacy of the noise is further amplified through iterative optimization using the PGD attack method.}
    \label{TBA}
\end{figure*}

In this paper, we focus on adversarial attacks targeting EHPS, a pivotal task in digital human generation. Our approach generates adversarial samples that expose vulnerabilities in state-of-the-art digital human models, demonstrating their susceptibility to even small, imperceptible perturbations.

\subsection{Tangible Attack (TBA) Framework}
Our objective is to investigate the robustness of digital human generation models. To this end, we introduce the \textbf{Tangible Attack (TBA)}, a universal attack specifically targeting EHPS, the core component of digital human generation. TBA is designed to generate adversarial samples capable of executing attacks across a wide range of digital human generation models. These adversarial samples induce significant deviations in the generated digital humans and can even produce anatomically implausible errors, particularly in the torso region. The overall framework is depicted in Fig. \ref{TBA}.

\paragraph{Dual Heterogeneous Noise Generator} 
We propose an innovative Dual Heterogeneous Noise Generator (DHNG) that combines a Variational Autoencoder (VAE) with ControlNet, specifically designed to generate highly effective adversarial \textbf{Noise} for targeting EHPS. 

First, \textbf{Noise 1 ($\beta^1 = \text{VAE}(\epsilon, \mathcal{I}^{c}; \theta_1$)} is generated purely from the image perspective. The VAE takes Gaussian noise $\epsilon$ as input and extracts visual features from the clean image using a feature extractor comprising two convolutional layers and an average pooling layer. These features are encoded together with $\epsilon$ to form a latent representation that captures both noise and image details, as described below:
\begin{gather}
    \hbar  = E(\epsilon), \quad \ell = \Gamma(\mathcal{I}^c), \\
    \mu = \mu_\phi (\hbar, \ell), \quad \sigma = \sigma_\phi (\hbar, \ell), \\
    z = \mu_\phi + \sigma_\phi \odot \zeta,
\end{gather}

where $\epsilon \sim \mathcal{N}(0,1)$, $E$ is the encoder in the VAE, $\hbar$ denotes the latent representation, and $\Gamma$ is the feature extractor. The features $\hbar$ and $\ell$ are fused and passed through the encoders $\mu_\phi$ and $\sigma_\phi$ to obtain the mean $\mu$ and variance $\sigma$ of the latent space. The latent variable $z$ is then sampled using reparameterization, where $\zeta \sim \mathcal{N}(0,1)$. Thereafter, the decoder $D$ transforms $z$ back into the data space to generate $\beta^1$:
\begin{equation}
    \beta^1 = D(z).
\end{equation}

Second, \textbf{Noise 2 ($\beta^2 = \text{ControlNet}(\epsilon, \mathcal{I}^{c}, T_f; \theta_2)$)} is generated from a text-image integration perspective by fine-tuning the Stable Diffusion model \cite{rombach2022high} within ControlNet \cite{zhang2023adding}. This allows for spatially controlled noise perturbations, which are sufficiently impactful to disrupt EHPS accuracy. Specifically, the noise image is encoded into the latent space $\chi$, and adversarial text embeddings $T_f$ and clean images $\mathcal{I}^c$ are introduced to generate perturbations with controlled effects. The noise addition and removal process follows a stepwise diffusion approach, with perturbations added at each time step $t$:
\begin{equation}
    \chi_t = \chi + \alpha_t \epsilon,
\end{equation}

where $\alpha_t$ controls the noise intensity at each step. ControlNet’s zero convolution module is utilized to regulate perturbations during early training, ensuring minimal disruption to the pre-trained model. By embedding the adversarial conditional input $T_f$, the model learns to generate adversarial perturbations, minimizing the error between predicted and actual noise:
\begin{equation}
    \mathcal{L}_{\mathrm{Con}} = \mathbb{E}_{\chi_0, t, T_t, T_f, \mathcal{I}^c, \epsilon } \left[ \|\epsilon   - \epsilon_{\theta_2}(\chi_t, t, T_t, T_f, \mathcal{I}^c) \|^2 \right].
\end{equation}

\paragraph{Noise Optimization}
For 3D parametric estimation, the EHPS model $\mathcal{F}$ outputs three types of parameters: pose $\delta \in \Omega^{53 \times 3}$, shape $\rho \in \Omega^{10}$, and expression $\varphi \in \Omega^{10}$. The goal of the attack is to perturb these parameters without considering the ground truth. We propose an adversarial loss function that quantifies the attack’s effectiveness via the negative average weighted distance of the parameters between the clean image $\mathcal{I}^c$ and its adversarial counterpart $\mathcal{I}^{adv} = \lambda \mathcal{I}^c + (1-\lambda) \beta$, where $\lambda = 0.5$:
\begin{equation}
    \mathcal{L}_{adv} = -\sum_{k \in \left\{ \delta, \rho, \varphi \right\}} \left \| \mathcal{F}(\mathcal{I}^{adv}) - \mathcal{F}(\mathcal{I}^{c}) \right \|^2.
\end{equation}

To prevent $\mathcal{I}^{adv}$ from deviating too far from $\mathcal{I}^{c}$, we also employ a mean square error (MSE) loss to maintain pixel-level consistency:
\begin{equation}
    \mathcal{L}_{\mathrm{MSE}} = \frac{1}{N}\sum_{i=1}^{N} \sum_{x,y} [\mathcal{I}_i^{adv}(x,y) - \mathcal{I}_i^{c}(x,y)]^2,
\end{equation}

where $x$ and $y$ refer to pixel coordinates.

Given the unique roles of VAE and ControlNet, we optimize them separately. The VAE is optimized using the following loss:
\begin{equation}
    \mathcal{L}_1 = \mathcal{L}_{\mathrm{MSE}} + \mathcal{L}_{adv}.
\end{equation}

For ControlNet, the optimization objective includes the adversarial loss and the ControlNet-specific loss:
\begin{equation}
    \mathcal{L}_2 = \mathcal{L}_{\mathrm{Con}} + \mathcal{L}_{\mathrm{MSE}} + \mathcal{L}_{adv}.
\end{equation}

After optimization, the final adversarial noise $\beta$ is obtained by combining $\beta^1$ and $\beta^2$: $\beta = \beta^1 + \beta^2$.

\begin{algorithm}[hb!]
    \caption{Generation of Adversarial Sample}
    \label{adv}
    \textbf{Input:} Random Gaussian Noise $\epsilon$, Original Image $\mathcal{I}^{c}$, Adversarial Text $T_f$\\
    \textbf{Output:} Adversarial Sample $\mathcal{I}^{adv}$ \\
    \textbf{Setting:} Epochs $N$ = 10, PGD Attack iterations $n$, Step size $\upsilon$, Perturbations threshold $\vartheta$, Mean square error loss $\mathcal{L}_{\mathrm{MSE}}$, Adversarial loss $\mathcal{L}_{adv}$, Diffusion loss for ControlNet $\mathcal{L}_{\mathrm{Con}}$, VAE model parameters $\theta_1$, ControlNet parameters $\theta_2$, SMPLer-X $\mathcal{F}$
    \begin{algorithmic}[1] %[1] enables line numbers
        \STATE Initialize parameters of $\theta_1$, $\theta_2$
        \FOR{$i$ = 1 to $N$}
            \STATE  $\beta^1$ = VAE($\epsilon$, $\mathcal{I}^{c}$; $\theta_1$)
            \STATE  $\epsilon  ^\ast $, $\beta^2$ = ControlNet($\epsilon$, $\mathcal{I}^{c}$, $T_f$; $\theta_2$)
            \STATE  $\beta$ = $\beta^1$ + $\beta^2$
            \STATE  $\mathcal{I}^{adv}$ = $\mathcal{I}^{c}$ + $\beta$
            \STATE  $\mathcal{L}_{1} = \mathcal{L}_{\mathrm{MSE}}(\mathcal{I}^{c}, \mathcal{I}^{adv}) + \mathcal{L}_{adv}(\mathcal{F}(\mathcal{I}^{c}), \mathcal{F}(\mathcal{I}^{adv}))$
            \STATE  $\mathcal{L}_{2} = \mathcal{L}_{\mathrm{Con}}(\epsilon, \epsilon  ^\ast ) + \mathcal{L}_{\mathrm{MSE}}(\mathcal{I}^{c}, \mathcal{I}^{adv}) + \mathcal{L}_{adv}(\mathcal{F}(\mathcal{I}^{c}), \mathcal{F}(\mathcal{I}^{adv}))$
            \STATE  \textbf{Update Parameters}
            \STATE  $\theta_1 = \theta_1 - \eta \nabla_{\theta_1}\mathcal{L}_1$
            \STATE $\theta_2 = \theta_2 - \eta \nabla_{\theta_2}\mathcal{L}_2$
            % Updating theta
        \ENDFOR
        \STATE \textbf{return} $\beta$ 
        \FOR{$j$ = 1 to $n$}
        \STATE $\beta$ = PGD($\beta$, $\upsilon$, $\vartheta$, $\mathcal{F}$, $\mathcal{I}^{c}$)
        \ENDFOR 
        \STATE \textbf{return} $\mathcal{I}^{adv} =\mathcal{I}^{c} + \beta$
    \end{algorithmic}
\end{algorithm}

\paragraph{Enhancement} 
To further amplify the destructive capability of the adversarial samples, we employ Projected Gradient Descent (PGD) \cite{madry2018towards}. We integrate gradient signals from both the noise and the SMPL-X model to guide the generation of more potent adversarial samples. Specifically, we initialize the noise with $\beta$ generated by DHNG and iteratively refine it by querying the SMPL-X model, using feedback from both the model’s gradients and the noise's gradients:

\begin{gather}
    \mathcal{I}_n^{adv} = \mathcal{I}^{c} + \beta_n, \\
    \beta_{n+1} = \beta_n + \upsilon \cdot \mathrm{sign}(\nabla_{\beta_n} \mathcal{L}(\mathcal{F}(\mathcal{I}_n^{adv}), \mathcal{F}(\mathcal{I}^{c}))), \\
    \beta_{n+1} = \beta_0 + \mathrm{CLIP}_{\left[ -\vartheta, \vartheta \right]}(\beta_{n+1} - \beta_0),
\end{gather}

where $n$ is the iteration number, $\upsilon$ is the step size, and $\vartheta$ constrains the perturbations within a threshold via clipping.

In summary, the process flow of the TBA framework for generating adversarial samples is presented in Algorithm \ref{adv}.

%===============================================
%===============================================
\section{Experiments}

In this section, we provide a detailed description of the experimental setup and present results on the 3DPW \cite{von2018recovering} and UBody \cite{lin2023one} datasets. The experiments were conducted in the following environment: Ubuntu 20.04.6 LTS, PyTorch framework, a single NVIDIA A100 GPU, and 40 GB of memory.

\subsection{Experimental Setup}

We begin by outlining the datasets, EHPS models, and evaluation metrics used in our experiments.

\noindent \textbf{Datasets.} To assess the robustness of EHPS models, we employ two widely used datasets: 3DPW and UBody. 
% These datasets serve as benchmarks to evaluate the impact of our adversarial attack on the accuracy of digital human generation.

\begin{itemize}
    \item \textbf{3DPW} \cite{von2018recovering} is a dataset featuring in-the-wild RGB video sequences annotated with 3D SMPL poses. It captures a broad range of movements performed by multiple actors in both indoor and outdoor settings, using a combination of single RGB cameras and inertial measurement units (IMUs) attached to the subjects.
    
    \item \textbf{UBody} \cite{lin2023one} is a large-scale dataset designed for upper-body mesh recovery tasks, bridging the gap between full-body reconstruction and real-world applications. It comprises 1,051K high-resolution images from 15 real-world scenarios, annotated with 2D whole-body keypoints, person bounding boxes, hand bounding boxes, and SMPL-X labels.
\end{itemize}

% \textbf{3D Poses in the Wild (3DPW)} \cite{von2018recovering} is a dataset consisting of in-the-wild RGB video sequences labeled with 3D SMPL poses. It includes a variety of movements performed by multiple actors in both indoor and outdoor environments, captured using a single RGB camera and inertial measurement units (IMUs) mounted on the subjects. This dataset is particularly valuable for evaluating the impact of our attack on the EHPS model, specifically in terms of its generation accuracy for major digital human body parts. \textbf{UBody} \cite{lin2023one} is a large-scale dataset of upper-body images designed to bridge the gap between full-body mesh recovery tasks and real-world applications. It includes data from 15 real-world scenarios, comprising 1,051K high-quality images annotated with 2D whole-body keypoints, person bounding boxes, hand bounding boxes, and SMPL-X labels. This dataset serves as a benchmark to evaluate the impact of our attack on the accuracy of the EHPS model in generating full-body digital human representations.

\noindent \textbf{EHPS Models.} We employ various state-of-the-art EHPS models as benchmarks for the black-box models to evaluate the attack efficacy of TBA. We apply Hand4Whole \cite{moon2022accurate}, OSX \cite{lin2023one}, and SMPLer-X \cite{cai2024smpler} as benchmarks in experiments. Specifically, SMPLer-X releases four foundation models, named ``SMPLer-X-M'', where M indicates the size of the ViT~\cite{dosovitskiy2020image} backbone (S, B, L, H).

\noindent \textbf{Evaluation Metrics.} We report the Mean Per Joint Position Error (MPJPE) and Mean Per-Vertex Position Error (MPVPE), which are used to assess 3D joint and mesh vertex positions, respectively, following existing EHPS methods \cite{moon2022accurate, lin2023one, cai2024smpler}. We also report the PA MPJPE and PA MPVPE, which further align rotation and scale. All metrics are reported in millimeters (mm).

% \noindent \textbf{Evaluation Metrics.}

\subsection{Attack Efficiency on EHPS Models}
To evaluate the attack efficacy of TBA in digital human generation, we use the adversarial samples generated by TBA to perform attacks on existing EHPS models. Notably, these adversarial samples are derived from the gradients of the SMPLer-X-H model. To simulate real-world attack scenarios, we avoid using any training data and generate adversarial samples exclusively from test data. Crucially, TBA does not require access to ground truth labels; instead, it focuses solely on inference results, aiming to amplify the discrepancy between the outputs of clean and adversarial samples. Once generated, the adversarial samples are used to attack state-of-the-art EHPS models without any prior knowledge of their internal structures or training data.

\begin{table}[htb]
  \centering
  \resizebox{0.49\textwidth}{!}{
  \begin{tabular}{lcc}
    \toprule[1.3pt]
Model & MPJPE (Body) $\downarrow$ \textit{(mm)} & PA MPJPE (Body) $\downarrow$ \textit{(mm)} \\
\hline
% HMR~\cite{kanazawa2018end} & 112.34 $|$   &   67.53 $|$  \\
Hand4Whole~\cite{moon2022accurate}  &   100.65 $|$ \textbf{113.96} ({\color{gray}13.22\%})   &  67.93 $|$ \textbf{74.29} ({\color{gray}8.94\%})  \\
OSX~\cite{lin2023one}   & 94.31 $|$ \textbf{103.02} ({\color{gray}9.24\%}) & 63.86  $|$ \textbf{68.84} ({\color{gray}7.80\%}) \\
SMPLer-X-S~\cite{cai2024smpler} & 82.67  $|$ \textbf{104.48} (\underline{{\color{gray}26.38\%}})  & 56.65 $|$ \textbf{67.55} (\underline{{\color{gray}19.24\%}}) \\
SMPLer-X-B~\cite{cai2024smpler}  & 79.46 $|$ \textbf{94.62} ({\color{gray}19.08\%})  & 52.62 $|$ \textbf{62.17} ({\color{gray}18.15\%})  \\
SMPLer-X-L~\cite{cai2024smpler}  & 75.85  $|$ \textbf{89.37} ({\color{gray}17.82\%})   & 50.67  $|$ \textbf{59.42} ({\color{gray}17.27\%})  \\
SMPLer-X-H~\cite{cai2024smpler}  & 75.01 $|$ \textbf{88.43} ({\color{gray}17.89\%})   & 50.57  $|$  \textbf{59.39} ({\color{gray}17.44\%})  \\
    \bottomrule[1.0pt]
  \end{tabular}}
    \caption{The attack effect of TBA on state-of-the-art EHPS models (Clean Samples $|$ Adversarial Samples) on the 3DPW dataset, with error growth rates marked in {\color{gray}gray}. The maximum error growth rate on each metric is \underline{underlined}.}
    \label{PW3D}
\end{table}

\begin{table*}[htb]
  \centering
  \resizebox{1.0\textwidth}{!}{
\begin{tabular}{lcccccccc}
\toprule[1.3pt]
\multirow{2}{*}{Model} & \multicolumn{3}{c}{PA MPVPE $\downarrow$ \textit{(mm)}}     & \multicolumn{3}{c}{MPVPE $\downarrow$ \textit{(mm)}} & \multicolumn{2}{c}{PA MPJPE $\downarrow$ \textit{(mm)}} \\
\cmidrule(r){2-4} \cmidrule(r){5-7} \cmidrule(r){8-9}
    & All   & Hands  & Face   & All  & Hands & Face        & Body      & Hands   \\
\hline
Hand4Whole~\cite{moon2022accurate}  & 43.16 $|$ \textbf{51.25} ({\color{gray}18.74\%})  & 8.32 $|$ \textbf{8.72} ({\color{gray}4.81\%}) &    2.98 $|$ \textbf{3.50} (\underline{{\color{gray}17.45\%}})  &  103.13 $|$ \textbf{127.88} ({\color{gray}24.00\%}) &  42.84 $|$ \textbf{44.12} ({\color{gray}2.99\%})  &   32.69 $|$ \textbf{32.71} ({\color{gray}0.06\%}) &   47.88 $|$ \textbf{63.52} (\underline{{\color{gray}32.66\%}})   &   8.48 $|$ \textbf{8.83} ({\color{gray}4.13\%})  \\
OSX~\cite{lin2023one}  & 40.85 $|$ \textbf{42.13} ({\color{gray}3.13\%}) & 9.38 $|$ \textbf{9.47} ({\color{gray}0.96\%})   & 3.33 $|$ \textbf{3.54} ({\color{gray}6.31\%})  & 86.37 $|$ \textbf{89.93} ({\color{gray}4.12\%}) & 42.73 $|$ \textbf{44.53} ({\color{gray}4.21\%}) & 25.71 $|$ \textbf{27.56} ({\color{gray}7.20\%}) & 49.25 $|$ \textbf{50.29} ({\color{gray}2.11\%}) & 9.63 $|$ \textbf{9.72} ({\color{gray}0.93\%})  \\
SMPLer-X-S~\cite{cai2024smpler}   & 32.28 $|$ \textbf{39.84} ({\color{gray}23.42\%}) & 10.05 $|$ \textbf{10.61} ({\color{gray}5.57\%}) & 2.83 $|$ \textbf{3.13} ({\color{gray}10.60\%})  & 57.23 $|$ \textbf{78.76} ({\color{gray}37.62\%}) & 42.80 $|$ \textbf{47.94} ({\color{gray}12.01\%}) & 22.26 $|$ \textbf{26.95} ({\color{gray}21.07\%}) & 37.38 $|$ \textbf{46.58} ({\color{gray}24.61\%}) & 10.27 $|$ \textbf{10.85} ({\color{gray}5.65\%}) \\
SMPLer-X-B~\cite{cai2024smpler}   & 28.95 $|$ \textbf{35.89} (\underline{{\color{gray}23.97\%}}) & 9.72 $|$ \textbf{9.81} ({\color{gray}0.93\%})   & 2.60 $|$ \textbf{2.87} ({\color{gray}10.38\%})  & 50.75 $|$ \textbf{70.22} ({\color{gray}38.36\%}) & 38.02 $|$ \textbf{42.94} ({\color{gray}12.94\%}) & 19.82 $|$ \textbf{24.71} (\underline{{\color{gray}24.67\%}}) & 33.43 $|$ \textbf{41.70} ({\color{gray}24.74\%}) & 9.91 $|$ \textbf{10.00} ({\color{gray}0.91\%}) \\
SMPLer-X-L~\cite{cai2024smpler}   & 25.66 $|$ \textbf{31.02} ({\color{gray}20.89\%}) & 8.98 $|$ \textbf{9.24} ({\color{gray}2.90\%})  & 2.39 $|$ \textbf{2.64} ({\color{gray}10.46\%})  & 43.26 $|$ \textbf{58.29} ({\color{gray}34.74\%}) & 33.11 $|$ \textbf{37.88} ({\color{gray}14.41\%}) & 17.54 $|$ \textbf{21.39} ({\color{gray}21.95\%}) & 30.14 $|$ \textbf{36.64} ({\color{gray}21.57\%}) & 9.16 $|$ \textbf{9.41} ({\color{gray}2.73\%})   \\
SMPLer-X-H~\cite{cai2024smpler}   & 24.64 $|$ \textbf{30.00} ({\color{gray}21.75\%})  & 8.47 $|$ \textbf{9.09} (\underline{{\color{gray}7.32\%}})    & 2.24 $|$ \textbf{2.43} ({\color{gray}8.48\%}) & 41.22 $|$ \textbf{58.13} (\underline{{\color{gray}41.02\%}}) & 30.88 $|$ \textbf{36.29} (\underline{{\color{gray}17.52\%}}) & 16.62 $|$ \textbf{21.19} ({\color{gray}27.50\%}) & 29.29 $|$ \textbf{35.36} ({\color{gray}20.72\%}) & 8.64 $|$ \textbf{9.25} (\underline{{\color{gray}7.06\%}})  \\
\bottomrule[1.0pt]
\end{tabular}}
\caption{The attack effect of TBA on state-of-the-art EHPS models (Clean Samples $|$ Adversarial Samples) on the UBody dataset, with error growth rates marked in {\color{gray}gray}. The maximum error growth rate on each metric is \underline{underlined}.}
\label{UBody}
\end{table*}

\paragraph{Quantitative Analysis}  
We first evaluate the effectiveness of TBA-generated adversarial samples through quantitative metrics on the 3DPW dataset, focusing on how these samples impact the body parts of digital humans produced by various EHPS models. The results, summarized in Table \ref{PW3D}, show that adversarial samples generated using the gradients of the SMPLer-X-H model lead to significant inaccuracies across all tested EHPS models. The minimum error growth rate observed is 7.80\%, with an average error growth rate of 16.04\%. Specifically, for the SMPLer-X-S model, the MPJPE increases from 82.67 mm to 104.48 mm, representing an error growth rate of 26.38\%. Notably, the error growth rates for the entire SMPLer-X series exceed 17.27\%, indicating that the attack is particularly effective across models sharing the same underlying architecture.

We also evaluate the adversarial impact of TBA on the UBody dataset, which includes whole-body, hand, and facial regions of digital humans. As shown in Table \ref{UBody}, the adversarial samples consistently reduce the accuracy of all EHPS models across all tasks. The average error growth rate reaches 14.69\%, with the MPVPE (All) for the SMPLer-X-H model increasing from 41.22 mm to 58.13 mm, representing an error growth rate of 41.02\%. Additionally, TBA achieves an average error growth rate of 31.17\% across MPVPE (All) metrics for all EHPS models, underscoring the general effectiveness of the attack across different tasks and body regions. Notably, the maximum error growth rates for PA MPVPE (Face) and PA MPJPE (Body) are observed in Hand4Whole, underscoring its significant attack power against other heterogeneous frameworks.

% \begin{figure}[t]
%     \centering
%     \includegraphics[width=0.45\textwidth]{3D_plot.png}
%     \caption{Comparison of clean and adversarial samples under PW3D: the gap between each pose parameter inferred by the SMPLer-X-H model and the ground truth parameters.}
%     \label{3d}
% \end{figure}

% \begin{figure}[htbp]
%     \centering
%     \begin{subfigure}{0.22\textwidth}
%         \centering
%         \includegraphics[width=\textwidth]{S.png} 
%         \caption{SMPLer-X-S}
%         \label{s}
%     \end{subfigure}
%     \hfill
%     \begin{subfigure}{0.22\textwidth}
%         \centering
%         \includegraphics[width=\textwidth]{B.png} 
%         \caption{SMPLer-X-B}
%         \label{b}
%     \end{subfigure}

%     % \vspace{1em} % 调整上下图片间距
%     \hfill

%     \begin{subfigure}{0.22\textwidth}
%         \centering
%         \includegraphics[width=\textwidth]{L.png} 
%         \caption{SMPLer-X-L}
%         \label{l}
%     \end{subfigure}
%     \hfill
%     \begin{subfigure}{0.22\textwidth}
%         \centering
%         \includegraphics[width=\textwidth]{H.png} 
%         \caption{SMPLer-X-H}
%         \label{h}
%     \end{subfigure}
%     % \caption{Comparison of clean and adversarial samples on PW3D: gaps between SMPLer-X inferred and ground truth parameters.}
%     \caption{Comparison of clean and adversarial samples under PW3D: the gap between each pose parameter inferred by SMPLer-X models and the ground truth parameters.}
%     \label{3d}
% \end{figure}

\begin{figure*}[htb]
    \centering
    \includegraphics[width=0.97\textwidth]{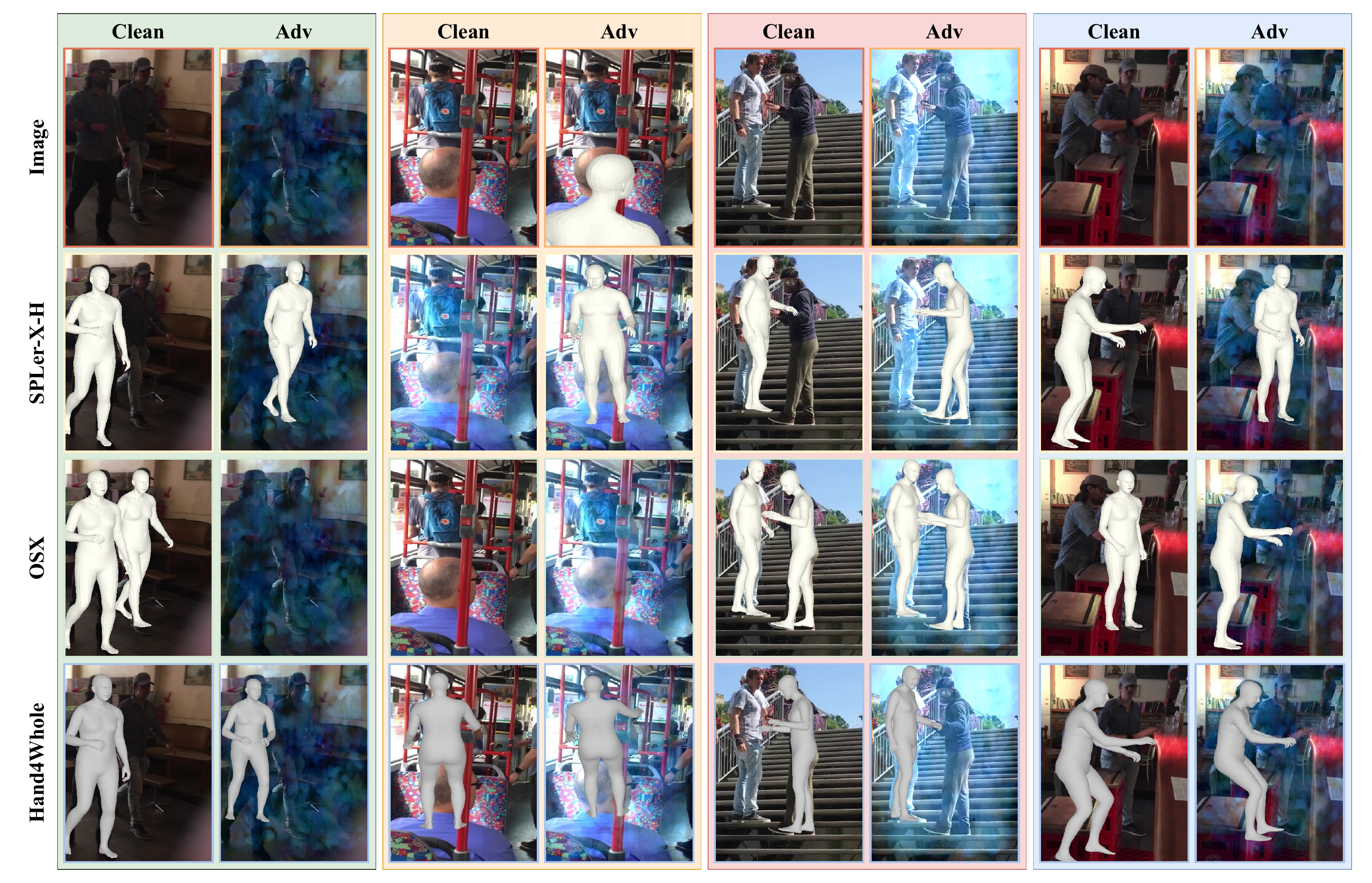}
    \caption{Visualizing \textbf{Clean} vs. \textbf{Adversarial (Adv)} samples of 3DPW for digital human generation under state-of-the-art EHPS models.}
    \label{vis_pw3d}
\end{figure*}

\begin{figure*}[htb]
    \centering
    \includegraphics[width=0.97\textwidth]{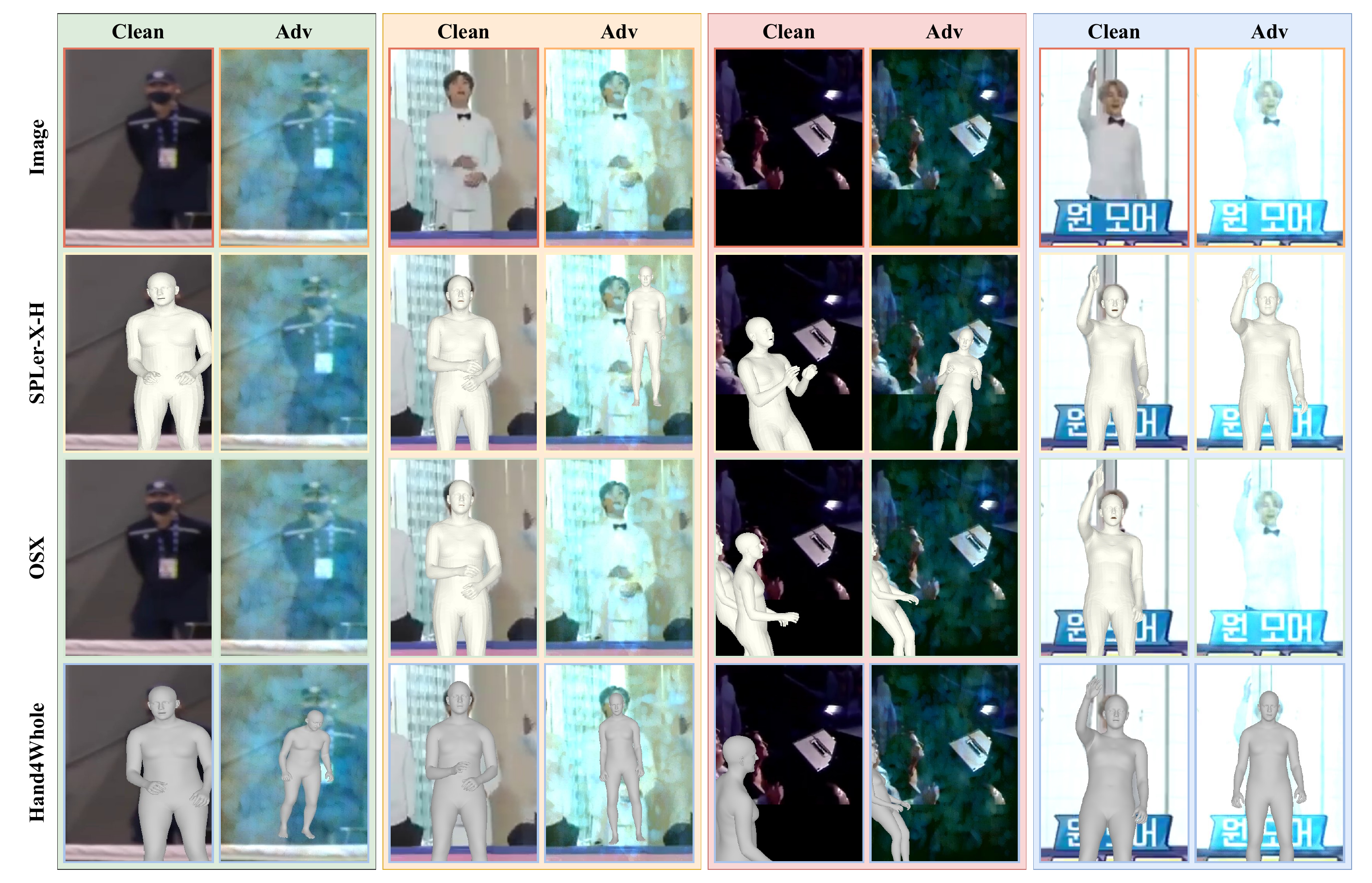}
    \caption{Visualizing \textbf{Clean} vs. \textbf{Adversarial (Adv)} samples of UBody for digital human generation under state-of-the-art EHPS models.}
    \label{vis_UBody}
\end{figure*}

\paragraph{Qualitative Analysis}  
To further illustrate TBA's attack performance, we conduct qualitative evaluations by visualizing the generated digital humans from both clean and adversarial samples across various scenarios. As shown in Fig.~\ref{vis_pw3d} and \ref{vis_UBody}, TBA causes a noticeable degradation in pose estimation accuracy, resulting in significant deviations in the generated digital humans, and in some cases, complete failures to generate a plausible human figure. These qualitative results visually demonstrate the attack's impact on the EHPS models' ability to accurately represent human poses.

\begin{figure}[htb]
    \centering
    \begin{subfigure}{0.23\textwidth}
        \centering
        \includegraphics[width=\textwidth]{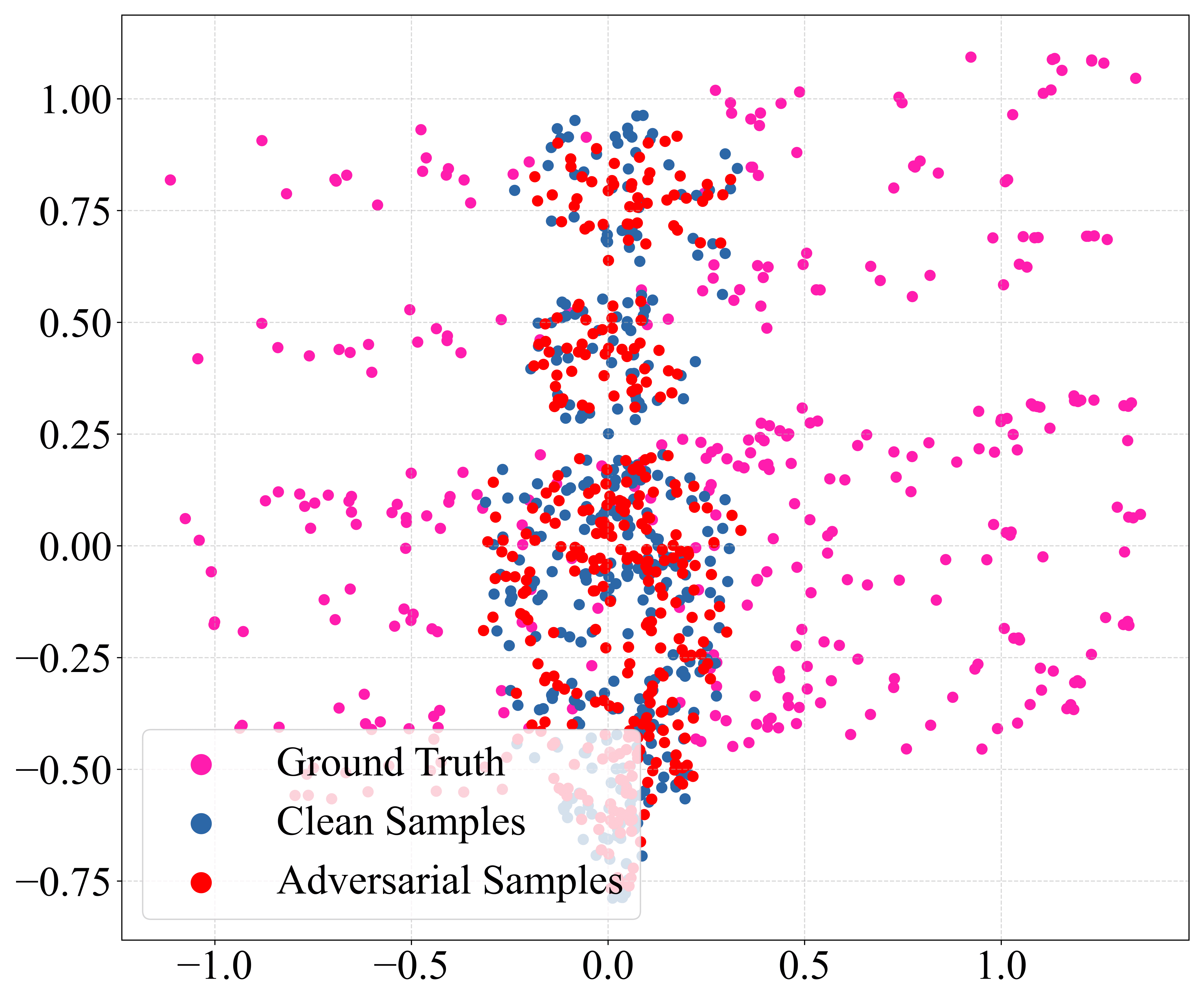} 
        \caption{SMPLer-X-S}
        \label{s}
    \end{subfigure}
    \hfill
    \begin{subfigure}{0.23\textwidth}
        \centering
        \includegraphics[width=\textwidth]{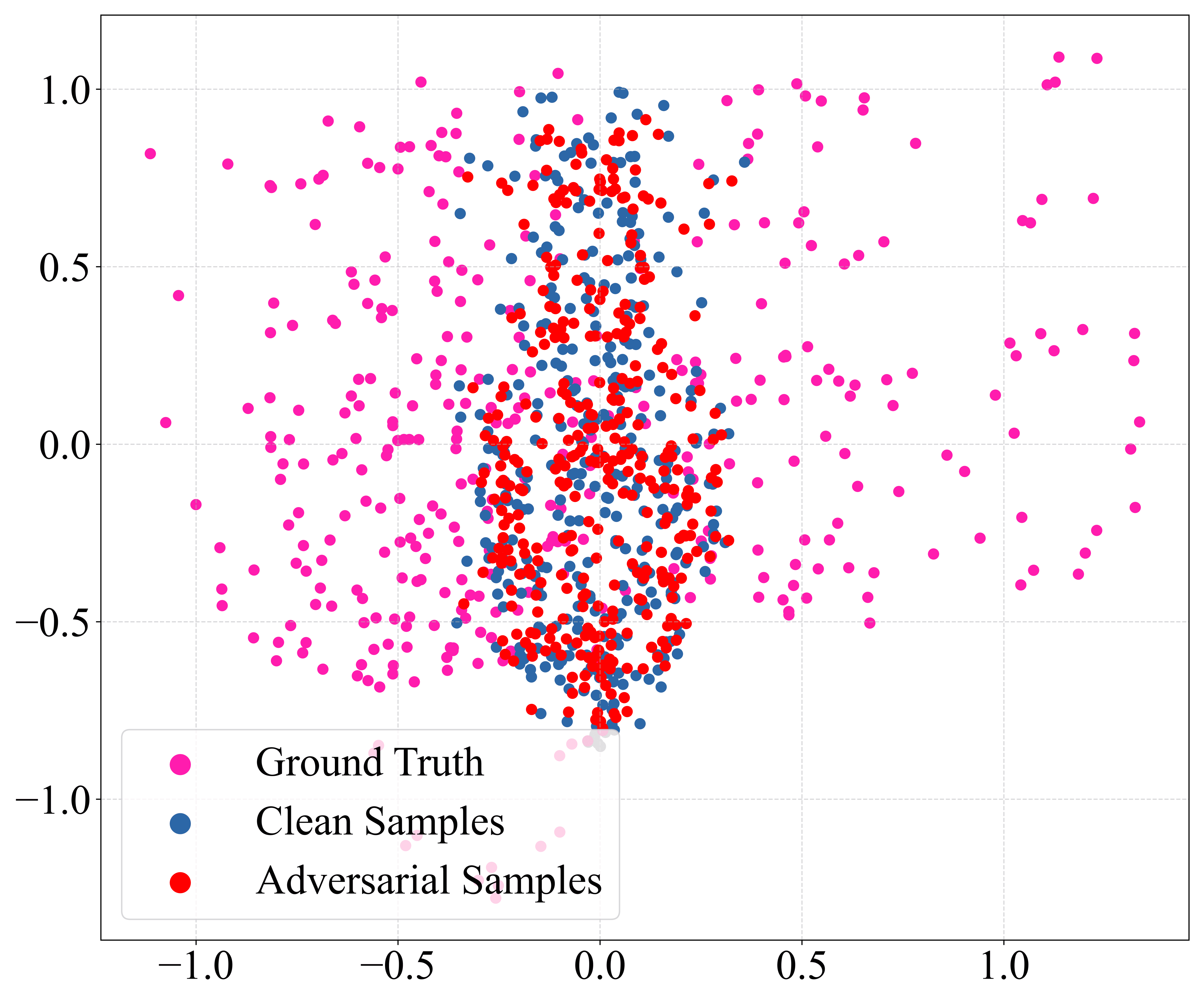} 
        \caption{SMPLer-X-B}
        \label{b}
    \end{subfigure}
    \hfill
    \begin{subfigure}{0.23\textwidth}
        \centering
        \includegraphics[width=\textwidth]{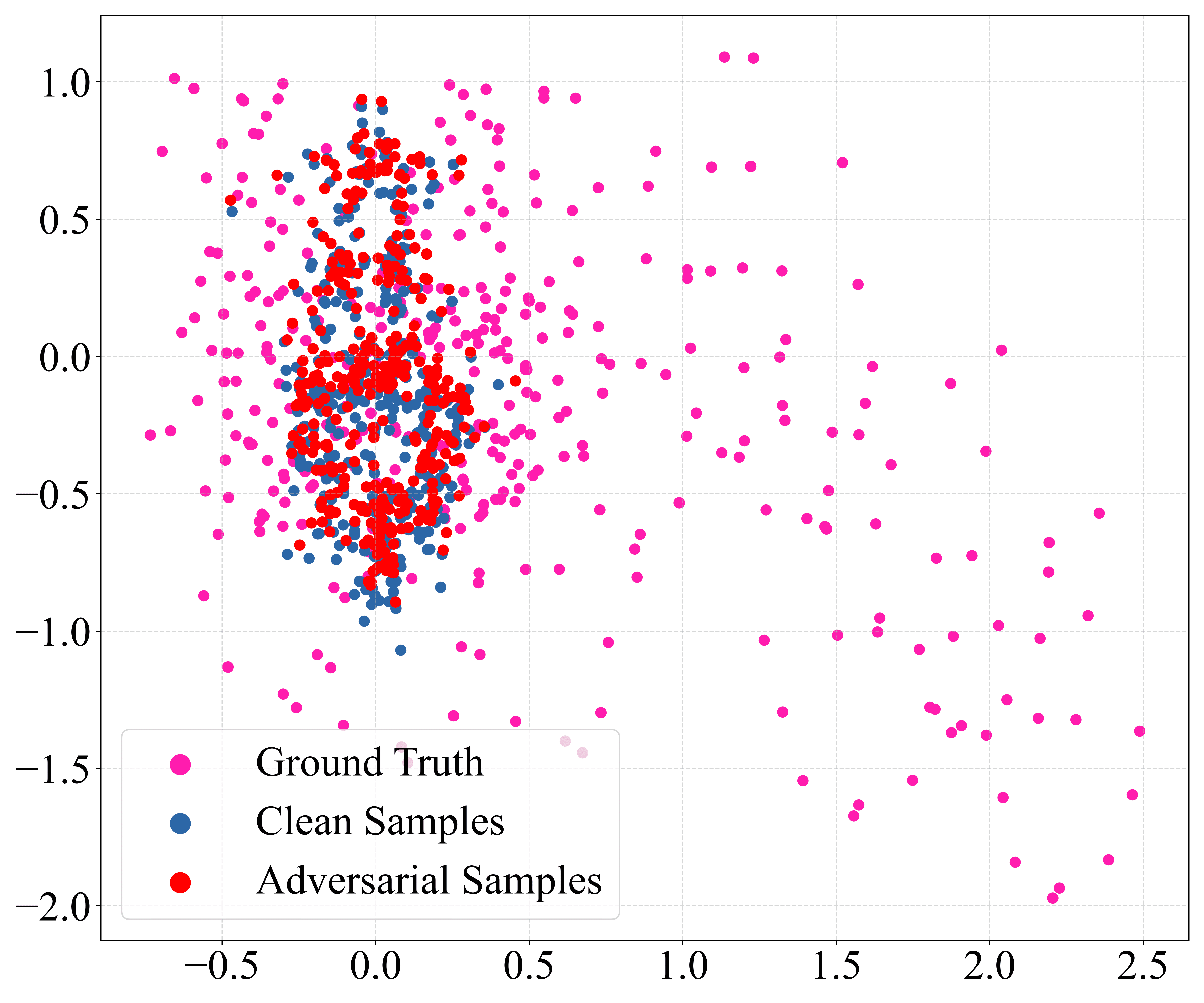} 
        \caption{SMPLer-X-L}
        \label{l}
    \end{subfigure}
    \hfill
    \begin{subfigure}{0.23\textwidth}
        \centering
        \includegraphics[width=\textwidth]{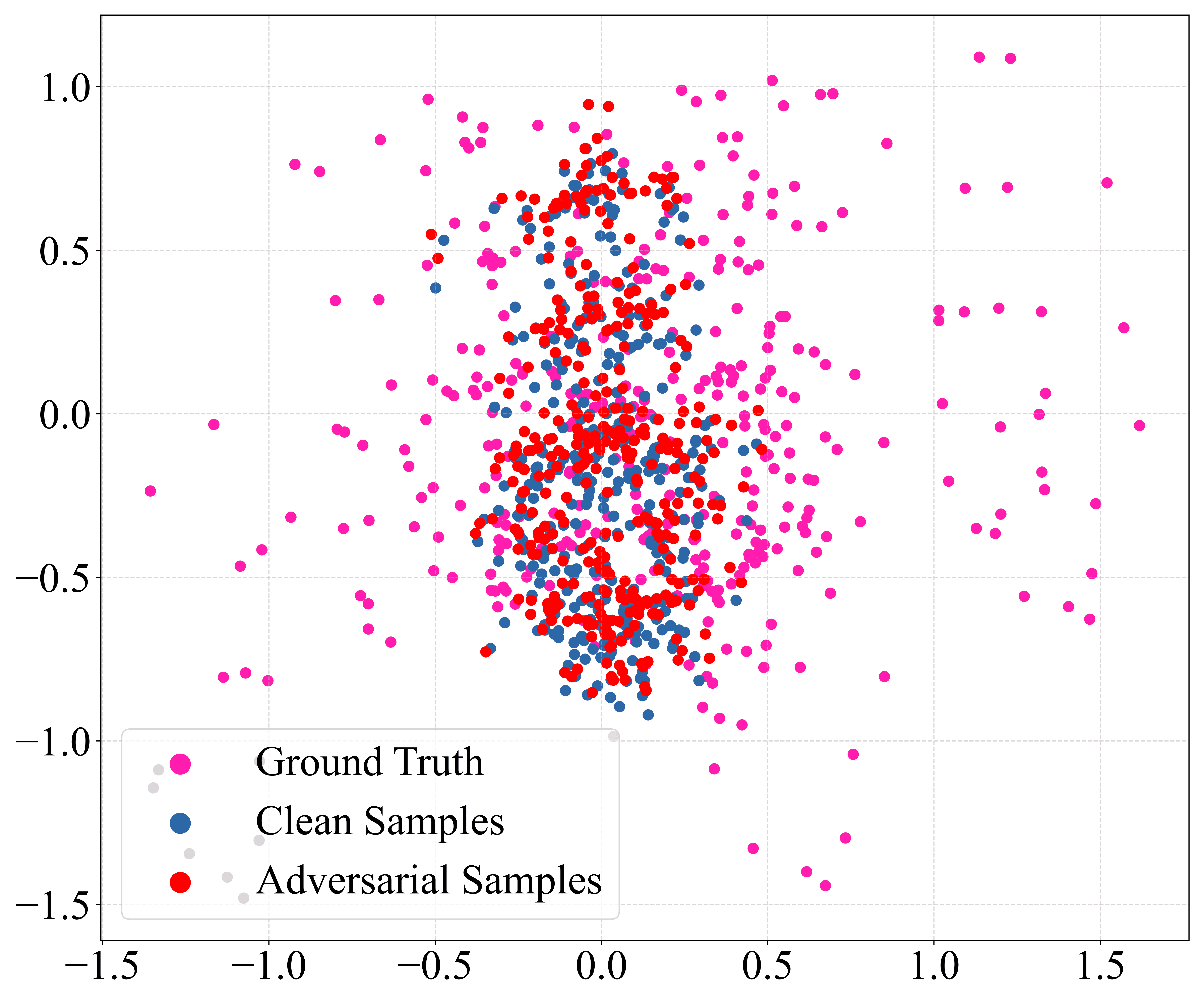} 
        \caption{SMPLer-X-H}
        \label{h}
    \end{subfigure}
    \caption{Comparison of clean and adversarial samples under PW3D: the gap between each pose parameter inferred by SMPLer-X models and the ground truth parameters.}
    \label{3d}
\end{figure}

Moreover, we analyze the data distribution of 3D joint points obtained from SMPLer-X’s inference for both clean and adversarial samples, comparing them with the ground truth joint points. As depicted in Fig.~\ref{3d}, adversarial samples generated by TBA exhibit deviations from the clean samples, with both distributions significantly diverging from the ground truth. This demonstrates the vulnerability of EHPS models to adversarial attacks, where the adversarial samples effectively disrupt the inference process, leading to significant pose estimation errors.

\subsection{Attack Efficiency on Different Scenarios}
To comprehensively evaluate the efficiency of TBA, we analyze its performance on the SMPLer-X model under both black-box and white-box attack scenarios using the 3DPW \cite{von2018recovering} and UBody \cite{lin2023one} datasets. Furthermore, we compare the attack effectiveness of TBA with the traditional adversarial attack method, FGSM \cite{goodfellow2014explaining}, addressing the significant gap in research on the robustness of digital human generation models.

\begin{table*}[htb]
  \centering
  \resizebox{0.8\textwidth}{!}{
  \begin{tabular}{lllllll}
  \toprule[1.3pt]
\multicolumn{1}{c}{\multirow{2}{*}{Model}} & \multicolumn{3}{c}{MPJPE(Body) $\downarrow$ \textit{(mm)}}          & \multicolumn{3}{c}{PA MPJPE(Body) $\downarrow$ \textit{(mm)}}     \\
\cmidrule(r){2-4} \cmidrule(r){5-7}
\multicolumn{1}{c}{} &
  \multicolumn{1}{c}{Clean} &
  \multicolumn{1}{c}{Adv (TBA)} &
  \multicolumn{1}{c}{Adv (FGSM)} &
  \multicolumn{1}{c}{Clean} &
  \multicolumn{1}{c}{Adv (TBA)} &
  \multicolumn{1}{c}{Adv (FGSM)} \\
\hline
SMPLer-X-S  & 82.67 & 104.48 (\underline{{\color{gray}26.38\%}}) & 84.53 ({\color{gray}2.25\%})  & 56.65 & 67.55 (\underline{{\color{gray}19.24\%}}) & 57.61 ({\color{gray}1.69\%}) \\
SMPLer-X-B  & 79.46 & 94.62 ({\color{gray}19.08\%})  & 82.51 ({\color{gray}3.84\%})  & 52.62 & 62.17 ({\color{gray}18.15\%}) & 54.52 ({\color{gray}3.61\%}) \\
SMPLer-X-L  & 75.85 & 89.37 ({\color{gray}17.82\%})  & 81.26 ({\color{gray}7.13\%})  & 50.67 & 59.42 ({\color{gray}17.27\%}) & 53.30 ({\color{gray}5.19\%}) \\
SMPLer-X-H  & 75.01 & 88.43 ({\color{gray}17.89\%})  & 91.23 ({\color{gray}21.62\%}) & 50.57 & 59.39 ({\color{gray}17.44\%}) & 58.30 ({\color{gray}15.29\%}) \\
    \bottomrule[1.0pt]
  \end{tabular}}
    % \caption{The black-box attack effect of TBA and FGSM on state-of-the-art EHPS models (SMPLer-X) on the 3DPW dataset, with error growth rates marked in {\color{gray}gray}. The maximum error growth rate on each metric is \underline{underlined}. Clean represents the clean samples, while Adv (TBA) and Adv (FGSM) represent the adversarial samples using TBA and FGSM, respectively.}
    \caption{The black-box attack effectiveness of TBA and FGSM on state-of-the-art EHPS models (SMPLer-X) is evaluated on the 3DPW dataset, with error growth rates highlighted in {\color{gray}gray}. The maximum error growth rate for each metric is \underline{underlined}. `Clean' refers to the original, unaltered samples, while `Adv (TBA)' and `Adv (FGSM)' represent the adversarial samples generated using TBA and FGSM, respectively.}
    \label{Black-box-attck_PW3D}
\end{table*}

\begin{table*}[htb]
    \centering
    \resizebox{1.0\textwidth}{!}{
    \centering
    \begin{tabular}{cccccccccc}
\toprule[1.3pt]
\multirow{3}{*}{Model} & \multicolumn{9}{c}{PA MPVPE $\downarrow$ \textit{(mm)}}                                                               \\
\cline{2-10}
                       & \multicolumn{3}{c}{All}      & \multicolumn{3}{c}{Hands}    & \multicolumn{3}{c}{Face}     \\
\cmidrule(r){2-4} \cmidrule(r){5-7} \cmidrule(r){8-10}
                       & Clean & Adv (TBA) & Adv (FGSM) & Clean & Adv (TBA) & Adv (FGSM) & Clean & Adv (TBA) & Adv (FGSM) \\
\hline
SMPLer-X-S & 32.28 & 39.84 ({\color{gray}23.43\%}) & 32.79 ({\color{gray}1.58\%})  & 10.05 & 10.61 ({\color{gray}5.57\%}) & 9.96 ({\color{gray}-0.90\%}) & 2.83 & 3.13 (\underline{{\color{gray}10.60\%}}) & 2.83 ({\color{gray}0.00\%}) \\
SMPLer-X-B & 28.95 & 35.89 ({\color{gray}23.97\%}) & 30.68 ({\color{gray}5.98\%})  & 9.72  & 9.81 ({\color{gray}0.93\%})  & 9.35 ({\color{gray}-3.81\%}) & 2.60 & 2.87 ({\color{gray}10.38\%}) & 2.64 ({\color{gray}1.54\%}) \\
SMPLer-X-L & 25.66 & 31.02 ({\color{gray}20.89\%}) & 29.39 ({\color{gray}14.54\%}) & 8.98  & 9.24 ({\color{gray}2.90\%})  & 8.69 ({\color{gray}-3.23\%}) & 2.39 & 2.64 ({\color{gray}10.46\%}) & 2.44 ({\color{gray}2.09\%}) \\
SMPLer-X-H & 24.64 & 30.00 ({\color{gray}21.75\%}) & 31.21 (\underline{{\color{gray}26.66\%}}) & 8.47  & 9.09 ({\color{gray}7.32\%})  & 9.12 (\underline{{\color{gray}7.67\%}})  & 2.24 & 2.43 ({\color{gray}8.48\%})  & 2.46 ({\color{gray}9.82\%}) \\
\bottomrule[1.0pt]
\end{tabular}}

    \vspace{1.5pt}
    
    \resizebox{1.0\textwidth}{!}{
    \centering
    \begin{tabular}{cccccccccc}
\toprule[1.3pt]
\multirow{3}{*}{Model} & \multicolumn{9}{c}{MPVPE $\downarrow$ \textit{(mm)}}                                                                  \\
\cline{2-10}
                       & \multicolumn{3}{c}{All}      & \multicolumn{3}{c}{Hands}    & \multicolumn{3}{c}{Face}     \\
\cmidrule(r){2-4} \cmidrule(r){5-7} \cmidrule(r){8-10}
                       & Clean & Adv (TBA) & Adv (FGSM) & Clean & Adv (TBA) & Adv (FGSM) & Clean & Adv (TBA) & Adv (FGSM) \\
\hline
SMPLer-X-S & 57.23 & 78.76 ({\color{gray}37.63\%}) & 59.51 ({\color{gray}3.98\%})  & 42.80 & 47.94 ({\color{gray}12.01\%}) & 43.29 ({\color{gray}1.14\%})  & 22.26 & 26.95 ({\color{gray}21.07\%}) & 22.51 ({\color{gray}1.12\%})  \\
SMPLer-X-B & 50.75 & 70.22 ({\color{gray}38.36\%}) & 58.05 ({\color{gray}14.38\%}) & 38.02 & 42.94 ({\color{gray}12.94\%}) & 39.34 ({\color{gray}3.47\%})  & 19.82 & 24.71 ({\color{gray}24.67\%}) & 20.35 ({\color{gray}2.67\%})  \\
SMPLer-X-L & 43.26 & 58.29 ({\color{gray}34.74\%}) & 56.52 ({\color{gray}30.65\%}) & 33.11 & 37.88 ({\color{gray}14.41\%}) & 35.94 ({\color{gray}8.55\%})  & 17.54 & 21.39 ({\color{gray}21.95\%}) & 19.63 ({\color{gray}11.92\%}) \\
SMPLer-X-H & 41.22 & 58.13 ({\color{gray}41.02\%}) & 64.27 (\underline{{\color{gray}55.92\%}}) & 30.88 & 36.29 ({\color{gray}17.52\%}) & 37.57 (\underline{{\color{gray}21.66\%}}) & 16.62 & 21.19 (\underline{{\color{gray}27.50\%}}) & 20.21 ({\color{gray}21.60\%}) \\
\bottomrule[1.0pt]
\end{tabular}}

\vspace{1.5pt}

\resizebox{0.8\textwidth}{!}{
\centering
\begin{tabular}{ccccccc}
\toprule[1.3pt]
\multirow{3}{*}{Model} & \multicolumn{6}{c}{PA MPJPE $\downarrow$ \textit{(mm)}}                                                         \\
\cline{2-7}
                       & \multicolumn{3}{c}{Body}                  & \multicolumn{3}{c}{Hands}                \\
\cmidrule(r){2-4}  \cmidrule(r){5-7}
                       & Clean & Adv (TBA)        & Adv (FGSM)       & Clean & Adv (TBA)       & Adv (FGSM)       \\
\hline
SMPLer-X-S             & 37.38 & 46.58 (\underline{{\color{gray}24.61\%}}) & 37.96 ({\color{gray}1.55\%})  & 10.27 & 10.85 ({\color{gray}5.64\%}) & 10.19 ({\color{gray}-0.78\%}) \\
SMPLer-X-B             & 33.43 & 41.70 ({\color{gray}24.74\%}) & 35.46 ({\color{gray}6.07\%})  & 9.91  & 10.00 ({\color{gray}0.91\%}) & 9.54 ({\color{gray}-3.73\%})  \\
SMPLer-X-L             & 30.14 & 36.64 ({\color{gray}21.57\%}) & 34.81 ({\color{gray}15.49\%}) & 9.16  & 9.41 ({\color{gray}2.73\%})  & 8.87 ({\color{gray}-3.17\%})  \\
SMPLer-X-H             & 29.29 & 35.36 ({\color{gray}20.72\%}) & 36.46 ({\color{gray}24.48\%}) & 8.64  & 9.25 ({\color{gray}7.06\%})  & 9.29 (\underline{{\color{gray}7.52\%}})  \\
\bottomrule[1.0pt]
\end{tabular} }
\caption{The black-box attack effectiveness of TBA and FGSM on state-of-the-art EHPS models (SMPLer-X) is evaluated on the UBody dataset, with error growth rates highlighted in {\color{gray}gray}. The maximum error growth rate for each metric is \underline{underlined}. `Clean' refers to the original, unaltered samples, while `Adv (TBA)' and `Adv (FGSM)' represent the adversarial samples generated using TBA and FGSM, respectively.}
    \label{Black-box-attck_UBody}
\end{table*}

\paragraph{Black-Box Attacks Scenario}
We primarily examine the attack effectiveness of adversarial samples generated using the FGSM and TBA attack methods on SMPLer-X in a black-box attack scenario. Specifically, adversarial samples are crafted using only the gradients of the SMPLer-X-H model, leveraging images outside the training set and without any prior knowledge of the training data. These adversarial samples are subsequently utilized to attack other digital human generation models without requiring any information about their internal architectures or training datasets.

\begin{table*}[htb]
  \centering
  \resizebox{0.8\textwidth}{!}{
  \begin{tabular}{lllllll}
  \toprule[1.3pt]
\multicolumn{1}{c}{\multirow{2}{*}{Model}} & \multicolumn{3}{c}{MPJPE(Body) $\downarrow$ \textit{(mm)}}          & \multicolumn{3}{c}{PA MPJPE(Body) $\downarrow$ \textit{(mm)}}     \\
\cmidrule(r){2-4} \cmidrule(r){5-7}
\multicolumn{1}{c}{} &
  \multicolumn{1}{c}{Clean} &
  \multicolumn{1}{c}{Adv (TBA)} &
  \multicolumn{1}{c}{Adv (FGSM)} &
  \multicolumn{1}{c}{Clean} &
  \multicolumn{1}{c}{Adv (TBA)} &
  \multicolumn{1}{c}{Adv (FGSM)} \\
\hline
SMPLer-X-S$^\ast$ & 82.67 & 108.15 (\underline{{\color{gray}30.82\%}}) & 103.38 ({\color{gray}25.05\%}) & 56.65 & 69.63 (\underline{{\color{gray}22.91\%}}) & 65.80 ({\color{gray}16.15\%}) \\
SMPLer-X-B$^\ast$ & 79.46 & 97.50 ({\color{gray}22.70\%}) & 94.00 ({\color{gray}18.30\%}) & 52.62 & 63.55 ({\color{gray}20.77\%}) & 59.92 ({\color{gray}13.87\%}) \\
SMPLer-X-L$^\ast$ & 75.85 & 91.63 ({\color{gray}20.80\%}) & 90.46 ({\color{gray}19.26\%}) & 50.67 & 60.54 ({\color{gray}19.48\%}) & 57.29 ({\color{gray}13.06\%}) \\
SMPLer-X-H$^\ast$ & 75.01 & 88.43 ({\color{gray}17.89\%}) & 91.23 ({\color{gray}21.62\%}) & 50.57 & 59.39 ({\color{gray}17.44\%}) & 58.30 ({\color{gray}15.29\%}) \\
    \bottomrule[1.0pt]
  \end{tabular}}
    \caption{The white-box attack effectiveness of TBA and FGSM on state-of-the-art EHPS models (SMPLer-X) is evaluated on the 3DPW dataset, with error growth rates highlighted in {\color{gray}gray}. The maximum error growth rate for each metric is \underline{underlined}. `Clean' refers to the original, unaltered samples, while `Adv (TBA)' and `Adv (FGSM)' represent the adversarial samples generated using TBA and FGSM, respectively. `SMPLer-X-M$^\ast$' represents the adversarial samples generated by the gradients of SMPLer-X-M.}
    \label{White-box-attack_PW3D}
\end{table*}

\begin{table*}[htb]
    \centering
    \resizebox{1.0\textwidth}{!}{
    \centering
    \begin{tabular}{cccccccccc}
\toprule[1.3pt]
\multirow{3}{*}{Model} & \multicolumn{9}{c}{PA MPVPE $\downarrow$ \textit{(mm)}}                                                               \\
\cline{2-10}
                       & \multicolumn{3}{c}{All}      & \multicolumn{3}{c}{Hands}    & \multicolumn{3}{c}{Face}     \\
\cmidrule(r){2-4} \cmidrule(r){5-7} \cmidrule(r){8-10}
                       & Clean & Adv (TBA) & Adv (FGSM) & Clean & Adv (TBA) & Adv (FGSM) & Clean & Adv (TBA) & Adv (FGSM) \\
\hline
SMPLer-X-S$^\ast$ & 32.28 & 39.52 ({\color{gray}22.43\%}) & 39.45 ({\color{gray}22.21\%}) & 10.05 & 10.96 (\underline{{\color{gray}9.05\%}}) & 10.01 ({\color{gray}-0.40\%}) & 2.83 & 3.17 ({\color{gray}12.01\%}) & 3.16 ({\color{gray}11.66\%}) \\
SMPLer-X-B$^\ast$ & 28.95 & 37.44 ({\color{gray}29.33\%}) & 35.56 ({\color{gray}22.83\%}) & 9.72  & 9.80 ({\color{gray}0.82\%})  & 9.68 ({\color{gray}-0.41\%})  & 2.60 & 2.90 ({\color{gray}11.54\%}) & 2.90 ({\color{gray}11.54\%}) \\
SMPLer-X-L$^\ast$ & 25.66 & 33.50 (\underline{{\color{gray}30.55\%}}) & 32.78 ({\color{gray}27.75\%}) & 8.98  & 9.43 ({\color{gray}5.01\%}) & 9.21 ({\color{gray}2.56\%}) & 2.39  & 2.69 (\underline{{\color{gray}12.55\%}}) & 2.56 ({\color{gray}7.11\%}) \\
SMPLer-X-H$^\ast$ & 24.64 & 30.00 ({\color{gray}21.75\%}) & 31.21 ({\color{gray}26.66\%}) & 8.47  & 9.09 ({\color{gray}7.32\%}) & 9.12 ({\color{gray}7.67\%}) & 2.24  & 2.43 ({\color{gray}8.48\%})  & 2.46 ({\color{gray}9.82\%}) \\
\bottomrule[1.0pt]
\end{tabular}}

    \vspace{1.5pt}
    
    \resizebox{1.0\textwidth}{!}{
    \centering
    \begin{tabular}{cccccccccc}
\toprule[1.3pt]
\multirow{3}{*}{Model} & \multicolumn{9}{c}{MPVPE $\downarrow$ \textit{(mm)}}                                                                  \\
\cline{2-10}
                       & \multicolumn{3}{c}{All}      & \multicolumn{3}{c}{Hands}    & \multicolumn{3}{c}{Face}     \\
\cmidrule(r){2-4} \cmidrule(r){5-7} \cmidrule(r){8-10}
                       & Clean & Adv (TBA) & Adv (FGSM) & Clean & Adv (TBA) & Adv (FGSM) & Clean & Adv (TBA) & Adv (FGSM) \\
\hline
SMPLer-X-S$^\ast$ & 57.23 & 81.48 ({\color{gray}42.37\%}) & 77.34 ({\color{gray}35.14\%}) & 42.80 & 49.46 ({\color{gray}15.56\%}) & 48.61 ({\color{gray}13.57\%}) & 22.26 & 28.72 ({\color{gray}29.02\%}) & 28.01 ({\color{gray}25.83\%}) \\
SMPLer-X-B$^\ast$ & 50.75 & 76.41 ({\color{gray}50.56\%}) & 70.90 ({\color{gray}39.70\%}) & 38.02 & 46.25 ({\color{gray}21.65\%}) & 43.12 ({\color{gray}13.41\%}) & 19.82 & 25.63 (\underline{{\color{gray}29.31\%}}) & 25.39 ({\color{gray}28.10\%}) \\
SMPLer-X-L$^\ast$ & 43.26 & 66.16 ({\color{gray}52.94\%}) & 62.43 ({\color{gray}44.31\%}) & 33.11 & 40.50 ({\color{gray}22.32\%}) & 39.32 ({\color{gray}18.76\%}) & 17.54 & 22.05 ({\color{gray}25.71\%}) & 21.93 ({\color{gray}25.03\%}) \\
SMPLer-X-H$^\ast$ & 41.22 & 58.13 ({\color{gray}41.02\%}) & 64.27 (\underline{{\color{gray}55.92\%}}) & 30.88 & 36.29 ({\color{gray}17.52\%}) & 37.58 (\underline{{\color{gray}21.70\%}}) & 16.62 & 21.19 ({\color{gray}27.50\%}) & 20.21 ({\color{gray}21.60\%}) \\
\bottomrule[1.0pt]
\end{tabular}}

\vspace{1.5pt}

\resizebox{0.8\textwidth}{!}{
\centering
\begin{tabular}{ccccccc}
\toprule[1.3pt]
\multirow{3}{*}{Model} & \multicolumn{6}{c}{PA MPJPE $\downarrow$ \textit{(mm)}}                                                         \\
\cline{2-7}
                       & \multicolumn{3}{c}{Body}                  & \multicolumn{3}{c}{Hands}                \\
\cmidrule(r){2-4}  \cmidrule(r){5-7}
                       & Clean & Adv (TBA)        & Adv (FGSM)       & Clean & Adv (TBA)       & Adv (FGSM)       \\
\hline
SMPLer-X-S$^\ast$ & 37.38 & 46.06 ({\color{gray}23.22\%}) & 45.57 ({\color{gray}21.91\%}) & 10.27 & 11.25 (\underline{{\color{gray}9.54\%}}) & 10.27 ({\color{gray}0.00\%}) \\
SMPLer-X-B$^\ast$ & 33.43 & 42.69 (\underline{{\color{gray}27.70\%}}) & 41.12 ({\color{gray}23.00\%}) & 9.91  & 10.04 ({\color{gray}1.31\%}) & 9.88 ({\color{gray}-0.30\%}) \\
SMPLer-X-L$^\ast$ & 30.14 & 39.10 ({\color{gray}29.73\%}) & 38.75 ({\color{gray}28.57\%}) & 9.16  & 9.60 ({\color{gray}4.80\%})  & 9.39 ({\color{gray}2.51\%})  \\
SMPLer-X-H$^\ast$ & 29.29 & 35.36 ({\color{gray}20.72\%}) & 36.46 ({\color{gray}24.48\%}) & 8.64  & 9.25 ({\color{gray}7.06\%})  &  9.29 ({\color{gray}7.52\%}) \\
\bottomrule[1.0pt]
\end{tabular} }

% \caption{The white-box attack effect of TBA and FGSM on state-of-the-art EHPS models (SMPLer-X) on the UBody dataset, with error growth rates marked in {\color{gray}gray}. The maximum error growth rate on each metric is \underline{underlined}.} 
\caption{The white-box attack effectiveness of TBA and FGSM on state-of-the-art EHPS models (SMPLer-X) is evaluated on the UBody dataset, with error growth rates highlighted in {\color{gray}gray}. The maximum error growth rate for each metric is \underline{underlined}. `Clean' refers to the original, unaltered samples, while `Adv (TBA)' and `Adv (FGSM)' represent the adversarial samples generated using TBA and FGSM, respectively. `SMPLer-X-M$^\ast$' represents the adversarial samples generated by the gradients of SMPLer-X-M.}
    \label{White-box-attack_UBody}
\end{table*}

\begin{figure*}[htb]
    \centering
    \includegraphics[width=1.0\textwidth]{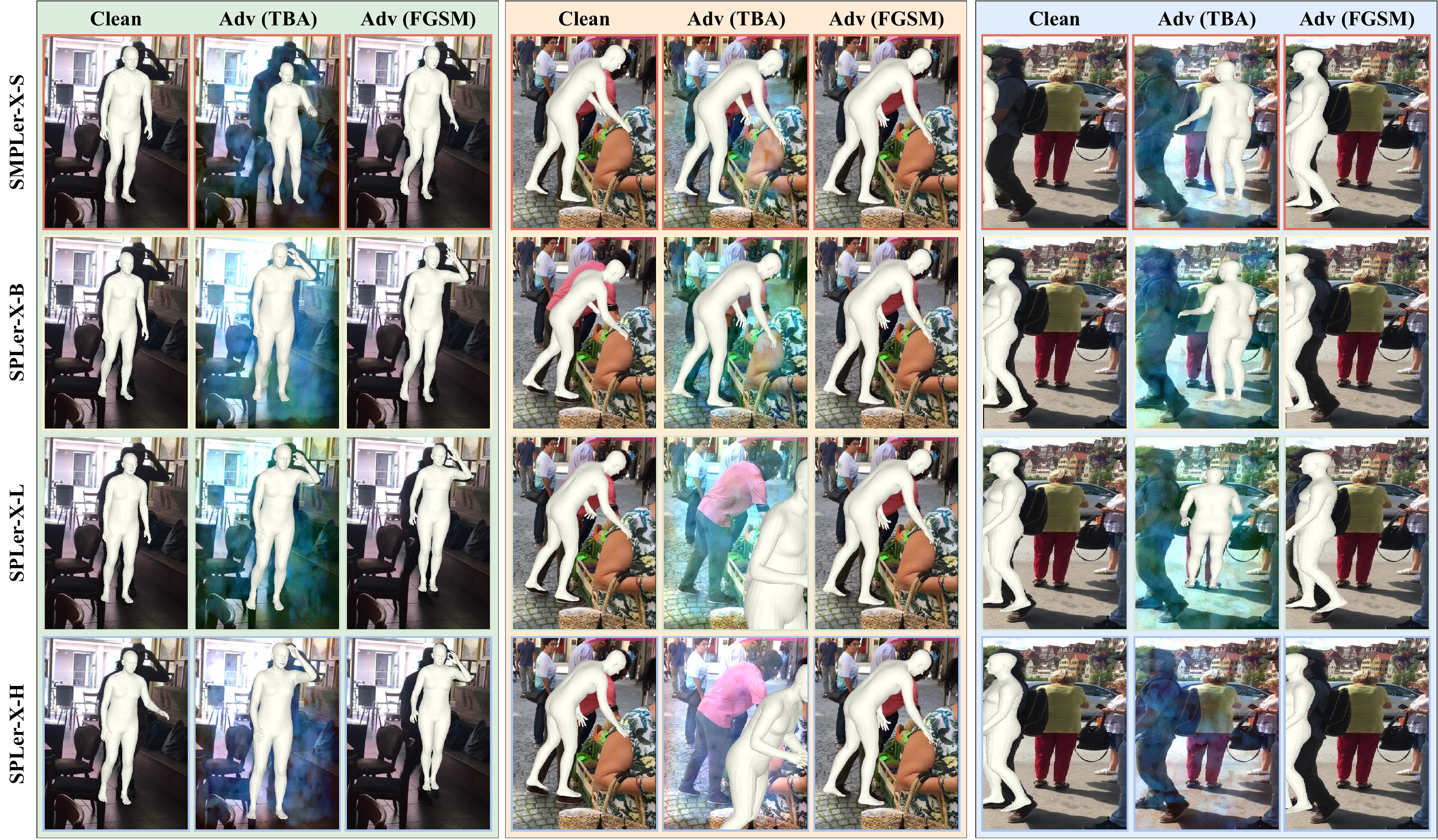}
    \caption{Visualizing \textbf{Clean} vs. \textbf{Adv (TBA)} VS. \textbf{Adv (FGSM)} samples of 3DPW for digital human generation under SMPLer-X models.}
    \label{vis_Sup_pw3d}
\end{figure*}

\begin{figure*}[htb]
    \centering
    \includegraphics[width=1.0\textwidth]{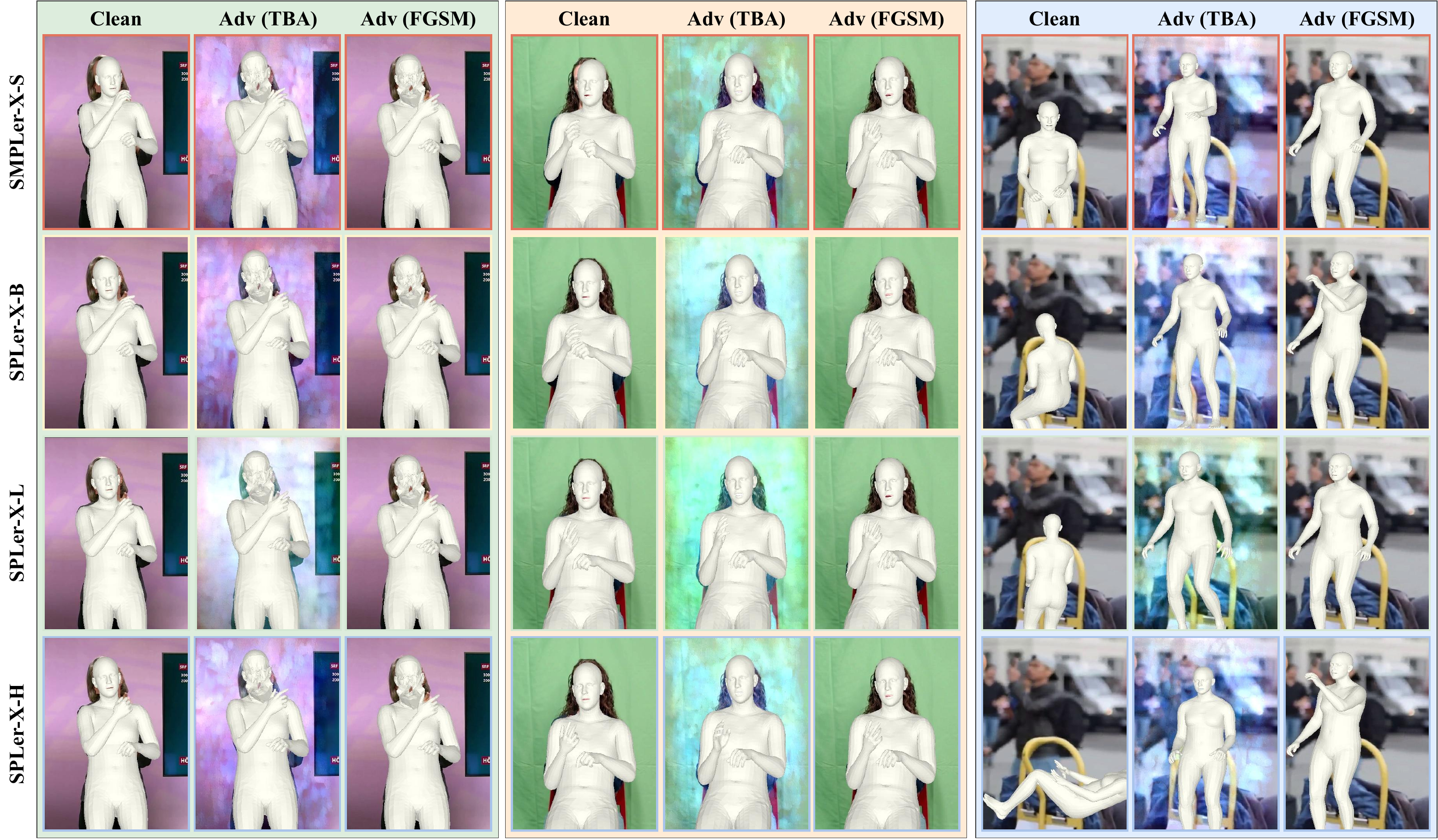}
    \caption{Visualizing \textbf{Clean} vs. \textbf{Adv (TBA)} VS. \textbf{Adv (FGSM)} samples of UBody for digital human generation under SMPLer-X models.}
    \label{vis_Sup_UBody}
\end{figure*}

As presented in Tables \ref{Black-box-attck_PW3D} and \ref{Black-box-attck_UBody}, TBA demonstrates more consistent attack performance compared to FGSM. While adversarial samples generated by FGSM achieve the highest error growth rate on several metrics, the error growth rates are notably non-uniform. For instance, although the maximum error growth rate of FGSM-generated adversarial samples reaches 55.92\% on the MPVPE (All) metric, the PA MPJPE (Hands) metric shows negative error growth rates for several models, including -0.78\% on SMPLer-X-S, -3.73\% on SMPLer-X-B, and -3.17\% on SMPLer-X-L. In contrast, adversarial samples generated by TBA effectively attack all digital human generation models, achieving a consistent average error growth rate exceeding 17.0\%.

\paragraph{White-Box Attacks Scenario}
In this section, we examine the effectiveness of adversarial samples generated using the FGSM and TBA methods in a white-box attack scenario targeting SMPLer-X. To achieve this, we specifically leverage the gradients of the four foundation models of SMPLer-X to produce adversarial samples. The generation process utilizes images from outside the training set, eliminating the need for prior knowledge about the training set's specifics. These adversarial samples are subsequently employed to attack the corresponding digital human generation models.

As illustrated in Tables \ref{White-box-attack_PW3D} and \ref{White-box-attack_UBody}, TBA demonstrates more robust and stable attack performance compared to FGSM. While adversarial samples generated using FGSM achieve the highest error growth rates in the MPVPE (All) and MPVPE (Hands) metrics, their performance is inconsistent, exhibiting instability and occasionally producing negative outcomes. In contrast, TBA-generated adversarial samples consistently achieve the maximum error growth rates across the remaining metrics, emphasizing their superior reliability and effectiveness. Additionally, we conduct visualization experiments, illustrated in Fig. \ref{vis_Sup_pw3d} and Fig. \ref{vis_Sup_UBody}, to further substantiate our findings.

\begin{table}[t]
  \centering
  \resizebox{0.47\textwidth}{!}{
  \begin{tabular}{lcc}
    \toprule[1.3pt]
Model & MPJPE (Body) $\downarrow$ \textit{(mm)} & PA MPJPE (Body) $\downarrow$ \textit{(mm)} \\
\hline
Original & 75.01 & 50.57 \\
VAE & 77.88 ({\color{gray}3.83\%}) & 54.19 ({\color{gray}7.16\%}) \\
ControlNet & 82.90 ({\color{gray}10.52\%}) & 54.62 ({\color{gray}8.01\%})\\
DHNG & 84.86 ({\color{gray}13.13\%}) & 55.36 ({\color{gray}9.47\%}) \\
PGD & 79.28 ({\color{gray}5.69\%}) & 54.30 ({\color{gray}7.38\%}) \\
DHNG + PGD &  86.46 ({\color{gray}15.26\%}) & 56.34 ({\color{gray}11.41\%})  \\
DHNG + Adv & 87.77 ({\color{gray}17.01\%}) & 59.92 ({\color{gray}18.49\%}) \\
TBA & 88.43 ({\color{gray}17.89\%}) & 59.39 ({\color{gray}17.44\%}) \\
    \bottomrule[1.0pt]
  \end{tabular}}
    \caption{The ablation study of TBA on SMPLer-X-H models on the 3DPW dataset, with error growth rates marked in {\color{gray}gray}. ``Adv'' represents the adversarial loss. }
    \label{abla}
\end{table}

\subsection{Ablation Study}
To rigorously evaluate the adversarial effectiveness of TBA, we first perform ablation studies across TBA's architectural framework to isolate and quantify the contribution of individual components to attack performance. Subsequently, we conduct robustness assessments on the EHPS model to comparatively validate TBA's superiority. Finally, a systematic investigation of ControlNet's textual conditions is implemented to investigate the impact of TBA's attack capabilities.

\paragraph{Different Component Contributions}
We conduct ablation experiments on the PW3D dataset to assess the contribution of different modules to the effectiveness of TBA in attacking the benchmark SMPLer-X-H model. As shown in Table \ref{abla}, when only the VAE or ControlNet is utilized as the noise generator, the error growth rates for the MPJPE and PA MPJPE metrics remain relatively low, indicating limited attack performance. In contrast, when the Dual Heterogeneous Noise Generator (DHNG) is employed, the error growth rate rises substantially to 13.13\% for MPJPE and 9.47\% for PA MPJPE, underscoring the pivotal role of DHNG in boosting the overall performance of TBA.

Moreover, combining DHNG with Projected Gradient Descent (PGD) further improves the attack, resulting in error growth rates of 15.26\% and 11.41\% for MPJPE and PA MPJPE, respectively, demonstrating that PGD enhances the effectiveness of DHNG. Similarly, the combination of DHNG with adversarial loss (``DHNG + Adv'') yields even higher error growth rates of 17.01\% for MPJPE and 18.49\% for PA MPJPE. Finally, the full TBA approach achieves error growth rates of 17.89\% for MPJPE and 17.44\% for PA MPJPE. These results indicate that the adversarial loss plays a crucial role in substantially improving the attack performance, while PGD contributes to performance enhancement, though to a lesser extent.

\begin{table*}[t]
  \centering
  \resizebox{0.80\textwidth}{!}{
\begin{tabular}{lcccccccc}
\toprule[1.3pt]
\multirow{2}{*}{Model} & \multicolumn{3}{c}{PA MPVPE $\downarrow$ \textit{(mm)}}     & \multicolumn{3}{c}{MPVPE $\downarrow$ \textit{(mm)}} & \multicolumn{2}{c}{PA MPJPE $\downarrow$ \textit{(mm)}} \\
\cmidrule(r){2-4} \cmidrule(r){5-7} \cmidrule(r){8-9}
    & All   & Hands  & Face   & All  & Hands & Face        & Body      & Hands   \\
\hline
Clean   & 24.64  & 8.47 & 2.24 & 41.22 & 30.88 & 16.62 & 29.29 & 8.64 \\
Random Noise $\uparrow$ & 24.66 & 8.47 & 2.24 & 41.26 & 30.88 & 16.62 & 29.31 & 8.64 \\

Random Noise $\uparrow \uparrow$ & 24.86 & 8.44 & 2.24 & 41.98 & 31.06 & 16.76 & 29.57 & 8.61\\

Random Noise $\uparrow \uparrow \uparrow$ & 25.34 & 8.43 & 2.25 & 44.29 & 31.44 & 17.07 & 30.22 & 8.59 \\

Random Noise $\uparrow \uparrow \uparrow \uparrow$ & 25.90 & 8.42 & 2.27 & 47.19 & 31.81 & 17.52 & 30.99 & 8.58 \\

Patch & 24.63 & 8.48 & 2.24 & 41.20 & 30.86 & 16.61 & 29.28 & 8.65\\

\textbf{TBA (Ours)} & \textbf{30.00} & \textbf{9.09} & \textbf{2.43} & \textbf{58.13} & \textbf{36.29} & \textbf{27.50} & \textbf{35.36} & \textbf{9.25} \\

\bottomrule[1.0pt]
\end{tabular}}
\caption{The attack effect of various attack methods on  SMPLer-X-H model on the UBody dataset. $\uparrow$ represents the random noise intensity.}
\label{roubust1}
\end{table*}

\paragraph{Roubustness of EHPS Model}
To systematically assess the robustness of the EHPS model, we attack the state-of-the-art SMPLer-X-H model with adversarial samples generated by four variable-intensity types of random noise and a white patch added in the middle of the image on the UBody dataset. As shown in Table \ref{roubust1}, performance metrics derived from adversarial samples generated by four types of random noise or the white patch exhibit statistically negligible deviations ($\leq$ 1\% ) from clean-sample results. This empirically confirms that ordinary noise fails to induce meaningful performance degradation in EHPS models, which is attributed to their robust generalization capacities enabled by large, diverse training dataset. Conversely, our proposed TBA can successfully compromise SMPLer-X-H, exposing critical vulnerabilities in current EHPS models against strategically designed adversarial samples.

\begin{table*}[t]
  \centering
  \resizebox{1.0\textwidth}{!}{
\begin{tabular}{lcccccccc}
\toprule[1.3pt]
\multirow{2}{*}{Text} & \multicolumn{3}{c}{PA MPVPE $\downarrow$ \textit{(mm)}}     & \multicolumn{3}{c}{MPVPE $\downarrow$ \textit{(mm)}} & \multicolumn{2}{c}{PA MPJPE $\downarrow$ \textit{(mm)}} \\
\cmidrule(r){2-4} \cmidrule(r){5-7} \cmidrule(r){8-9}
    & All   & Hands  & Face   & All  & Hands & Face        & Body      & Hands   \\
\hline
None   & 24.64 $|$ \textbf{30.46} (\underline{{\color{gray}23.62\%}})  & 8.47 $|$ \textbf{8.78} ({\color{gray}3.66\%})    & 2.24 $|$ \textbf{2.52} (\underline{{\color{gray}12.50\%}}) & 41.22 $|$ \textbf{57.98} ({\color{gray}40.66\%}) & 30.88 $|$ \textbf{36.20} ({\color{gray}17.23\%}) & 16.62 $|$ \textbf{21.03} ({\color{gray}26.53\%}) & 29.29 $|$ \textbf{34.86} ({\color{gray}19.02\%}) & 8.64 $|$ \textbf{9.16} ({\color{gray}6.02\%}) \\

Text1   & 24.64 $|$ \textbf{27.93} ({\color{gray}13.35\%})  & 8.47 $|$ \textbf{8.68} ({\color{gray}2.48\%})    & 2.24 $|$ \textbf{2.36} ({\color{gray}5.36\%}) & 41.22 $|$ \textbf{53.77} ({\color{gray}30.44\%}) & 30.88 $|$ \textbf{33.84} ({\color{gray}9.59\%}) & 16.62 $|$ \textbf{20.13} ({\color{gray}21.12\%}) & 29.29 $|$ \textbf{32.86} ({\color{gray}12.19\%}) & 8.64 $|$ \textbf{9.01} ({\color{gray}4.28\%}) \\

Text2   & 24.64 $|$ \textbf{27.16} ({\color{gray}10.22\%})  & 8.47 $|$ \textbf{8.66} ({\color{gray}2.24\%})    & 2.24 $|$ \textbf{2.32} ({\color{gray}3.57\%}) & 41.22 $|$ \textbf{53.02} ({\color{gray}28.62\%}) & 30.88 $|$ \textbf{32.97} ({\color{gray}6.77\%}) & 16.62 $|$ \textbf{19.44} ({\color{gray}16.97\%}) & 29.29 $|$ \textbf{32.07} ({\color{gray}9.49\%}) & 8.64 $|$ \textbf{8.92} ({\color{gray}3.24\%}) \\

Text3   & 24.64 $|$ \textbf{26.46} ({\color{gray}7.39\%})  & 8.47 $|$ \textbf{8.63} ({\color{gray}1.89\%})    & 2.24 $|$ \textbf{2.34} ({\color{gray}4.46\%}) & 41.22 $|$ \textbf{52.42} ({\color{gray}27.17\%}) & 30.88 $|$ \textbf{32.05} ({\color{gray}3.79\%}) & 16.62 $|$ \textbf{18.92} ({\color{gray}13.84\%}) & 29.29 $|$ \textbf{31.73} ({\color{gray}8.33\%}) & 8.64 $|$ \textbf{8.73} ({\color{gray}1.04\%}) \\

Text4 (Ours)  & 24.64 $|$ \textbf{30.00} ({\color{gray}21.75\%})  & 8.47 $|$ \textbf{9.09} (\underline{{\color{gray}7.32\%}})    & 2.24 $|$ \textbf{2.43} ({\color{gray}8.48\%}) & 41.22 $|$ \textbf{58.13} (\underline{{\color{gray}41.02\%}}) & 30.88 $|$ \textbf{36.29} (\underline{{\color{gray}17.52\%}}) & 16.62 $|$ \textbf{21.19} (\underline{{\color{gray}27.50\%}}) & 29.29 $|$ \textbf{35.36} (\underline{{\color{gray}20.72\%}}) & 8.64 $|$ \textbf{9.25} (\underline{{\color{gray}7.06\%}})  \\

\bottomrule[1.0pt]
\end{tabular}}
\caption{The attack effect of various attack methods on  SMPLer-X-H model (Clean Samples $|$ Adversarial Samples) on the UBody dataset, with error growth rates marked in {\color{gray}gray}. The maximum error growth rate on each metric is \underline{underlined}.}
\label{roubust2}
\end{table*}

\paragraph{Impact of Text embedding in Controlnet}
We investigate the impact of diverse textual conditions on TBA attack efficacy with varying ChatGPT-generated \cite{brown2020language} text prompts in ControlNet using the UBody dataset. The following four text conditions are evaluated:
\begin{itemize}
    \item Text1: Generate a distorted grid overlay with chaotic lines and unpredictable color patterns.

    \item Text2: A chaotic urban environment filled with blurry holograms and glitching lights.

    \item Text3: A holographic projection overlaying abstract fractal noise onto a humanoid figure.

    \item Text4: Generate an imperceptible noise overlay applied to the torso of a human pose. The noise should be invisible to the naked eye but designed to subtly distort or degrade the accuracy of human pose estimation models. Focus on embedding the noise naturally into the fabric and texture of the torso, ensuring that the person’s body and posture remain visually unchanged while embedding adversarial perturbations.
    
\end{itemize}

As shown in Table \ref{roubust2}, textual variations significantly affect TBA attack performance, with ``Text4'' demonstrating superior effectiveness. Specifically, ``Text4'' induces a 41\% increase in error rate for the MPVPE (All) metric. Other experiments adopt ``Text4'' as ControlNet's default textual conditioning parameter.

%==================================================
\section{Conclusions}
% We propose, for the first time, the Tangible Attack (TBA) method targeting expressive human pose and shape estimation (EHPS), thereby uncovering potential security risks and raising awareness within the research community. The core innovation of TBA lies in its dual heterogeneous noise generator, which leverages both VAE and ControlNet to craft specialized noise. A novel adversarial loss is introduced to guide the noise in degrading the accuracy of each pose parameter estimated by EHPS models. Furthermore, the adversarial samples generated by TBA are used to perform black-box attacks on arbitrary EHPS models, without any knowledge of their internal structures. Experimental results demonstrate the effectiveness of TBA, revealing that EHPS models are vulnerable to carefully crafted adversarial samples.
In this work, we introduce Tangible Attack (TBA), the first adversarial attack method specifically designed to target expressive human pose and shape estimation (EHPS). This novel approach sheds light on previously underexplored security vulnerabilities in EHPS models and calls for heightened attention from the broader research community regarding adversarial robustness in digital human modeling. The core innovation of TBA lies in its dual heterogeneous noise generator, which combines a Variational Autoencoder (VAE) and ControlNet to craft specialized adversarial perturbations. A novel adversarial loss is introduced to guide these perturbations in explicitly maximizing the deviation of each pose parameter predicted by EHPS models. Furthermore, the adversarial samples generated by TBA are used to perform black-box attacks on arbitrary EHPS models, without any knowledge of their internal structures. This generality significantly enhances the practicality and transferability of the attack in real-world applications. Extensive experimental results demonstrate the effectiveness of TBA, revealing that EHPS models are vulnerable to carefully crafted adversarial samples.

\noindent \textbf{Limitations \& Future Work.} 
% Our approach's limitation is that the adversarial samples we generate are visually perceptible, making them relatively easy to detect. Future work should focus on creating imperceptible adversarial samples for digital human generation. This research aims to inspire robust algorithm development and foster collaboration to enhance security.
% One limitation of our current approach is that the adversarial samples we generate are visually perceptible, making them relatively easy to detect. A key direction for future work is the development of imperceptible adversarial samples for digital human generation. We hope this research encourages further exploration of robust algorithms for digital human generation and fosters collaborative efforts to improve security within the community.
Despite its effectiveness, the TBA has a notable limitation: the generated adversarial samples often exhibit visually perceptible artifacts, which may limit their stealth in practical scenarios. A key direction for future work is the development of imperceptible adversarial samples for digital human generation. In addition, we plan to explore adaptive defense mechanisms and certifiably robust EHPS models that can detect or withstand such attacks.  Ultimately, we hope that this work serves as a catalyst for deeper investigation into the security and reliability of digital human representations and inspires the development of robust, interpretable, and secure EHPS systems. As digital humans become increasingly prevalent in applications ranging from AR/VR to human-computer interaction and virtual content creation, ensuring their integrity and safety is of paramount importance.

{
    \small
    \bibliographystyle{ieeenat_fullname}
    \bibliography{main}
}

% WARNING: do not forget to delete the supplementary pages from your submission 
% \input{sec/X_suppl}

\end{document}